\documentclass[letterpaper]{article} 
\usepackage{aaai23}  
\usepackage{times}  
\usepackage{helvet}  
\usepackage{courier}  
\usepackage[hyphens]{url}  
\usepackage{graphicx} 
\usepackage{natbib}  
\usepackage{caption} 
\usepackage{algorithm}
\usepackage{algorithmic}

\usepackage{newfloat}
\usepackage{listings}
\DeclareCaptionStyle{ruled}{labelfont=normalfont,labelsep=colon,strut=off} 
\floatstyle{ruled}
\newfloat{listing}{tb}{lst}{}
\floatname{listing}{Listing}
\usepackage{amsmath}
\usepackage{amsfonts}
\usepackage{booktabs}
\usepackage{pifont}
\newcommand{\cmark}{\ding{51}}%
\newcommand{\xmark}{\ding{55}}%

\usepackage{tikz}
\newcommand*\circled[1]{\tikz[baseline=(char.base)]{\node[shape=circle,draw,inner sep=1.5pt] (char) {#1};}}

\title{Confidence Contours: Uncertainty-Aware Annotation for Medical \\ Semantic Segmentation}
\author{
    Andre Ye, Quan Ze Chen, Amy Zhang
}
\affiliations{
    University of Washington \\
    
    andreye@uw.edu, cqz@cs.washington.edu, axz@cs.uw.edu
    
}

\begin{document}

\maketitle

\begin{abstract}
    Medical image segmentation modeling is a high-stakes task where understanding of uncertainty is crucial for addressing visual ambiguity. Prior work has developed segmentation models utilizing probabilistic or generative mechanisms to infer uncertainty from labels where annotators draw a singular boundary. However, as these annotations cannot represent an individual annotator's uncertainty, models trained on them produce uncertainty maps that are difficult to interpret. We propose a novel segmentation representation, Confidence Contours, which uses high- and low-confidence ``contours’’ to capture uncertainty directly, and develop a novel annotation system for collecting contours. We conduct an evaluation on the Lung Image Dataset Consortium (LIDC) and a synthetic dataset. 
    From an annotation study with 30 participants, results show that Confidence Contours provide high representative capacity without considerably higher annotator effort.
    We also find that general-purpose segmentation models can learn Confidence Contours at the same performance level as standard singular annotations. 
    Finally, from interviews with 5 medical experts, we find that Confidence Contour maps are more interpretable than Bayesian maps due to representation of structural uncertainty.
\end{abstract}

\section{Introduction}



Increasingly sophisticated general segmentation models such as U-Net~\cite{Ronneberger2015UNetCN} and DeepLab~\cite{Chen2016DeepLabSI} have become widely adopted for medical imaging problems~\cite{Lei2020MedicalIS}.
Despite their high expressive power and adaptability, these models often fail to represent contextual uncertainty in medical images, as evidenced by poorly visible structure borders, abnormally shaped structures, and other ambiguous features~\cite{Armato2011TheLI,Menze2015TheMB}. 
Models that fail to provide accurate uncertainty information can impede human users' ability to assess correctness for downstream decision-making~\cite{Chen2021ExplainableMI,gordon2021-deconvoltion}.
To address the shortcomings of conventional segmentation models, a growing body of work uses elicitive, multi-candidate, and generative methods to train uncertainty-aware models to produce continuous-valued uncertainty maps or to generate ensembles of candidate segmentations~\cite{Gawlikowski2021ASO}.
These models infer uncertainty from conventional \textit{singular} annotations for images, where each annotator draws a single boundary around the positive-class region. 



\begin{figure}
    \centering
    \setlength\tabcolsep{1.5pt}
    \begin{tabular}{lll}
        \includegraphics[width=2.6cm, height=2.6cm]{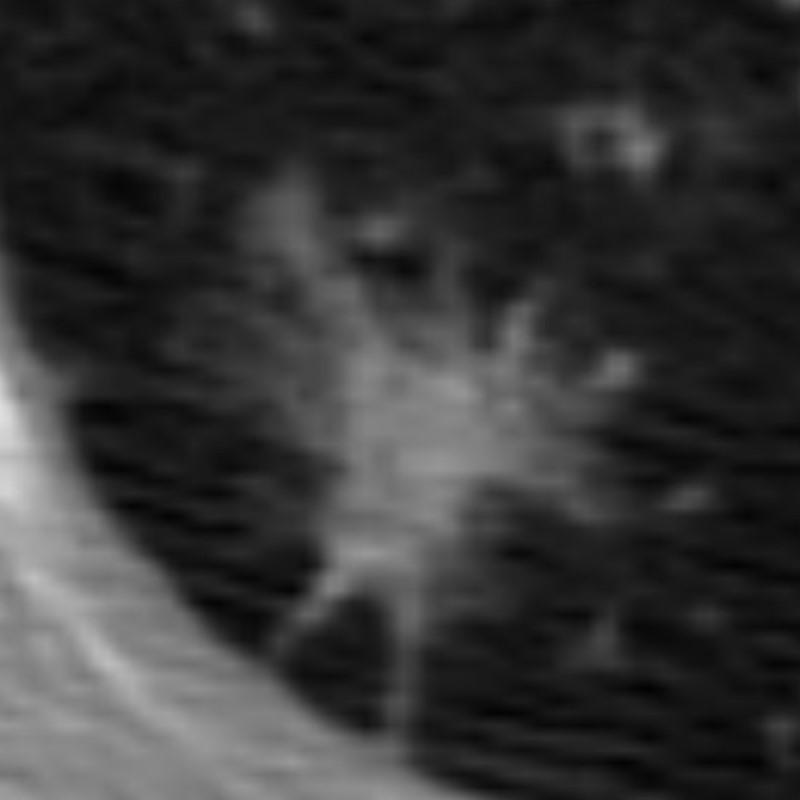} & 
        \includegraphics[width=2.6cm, height=2.6cm]{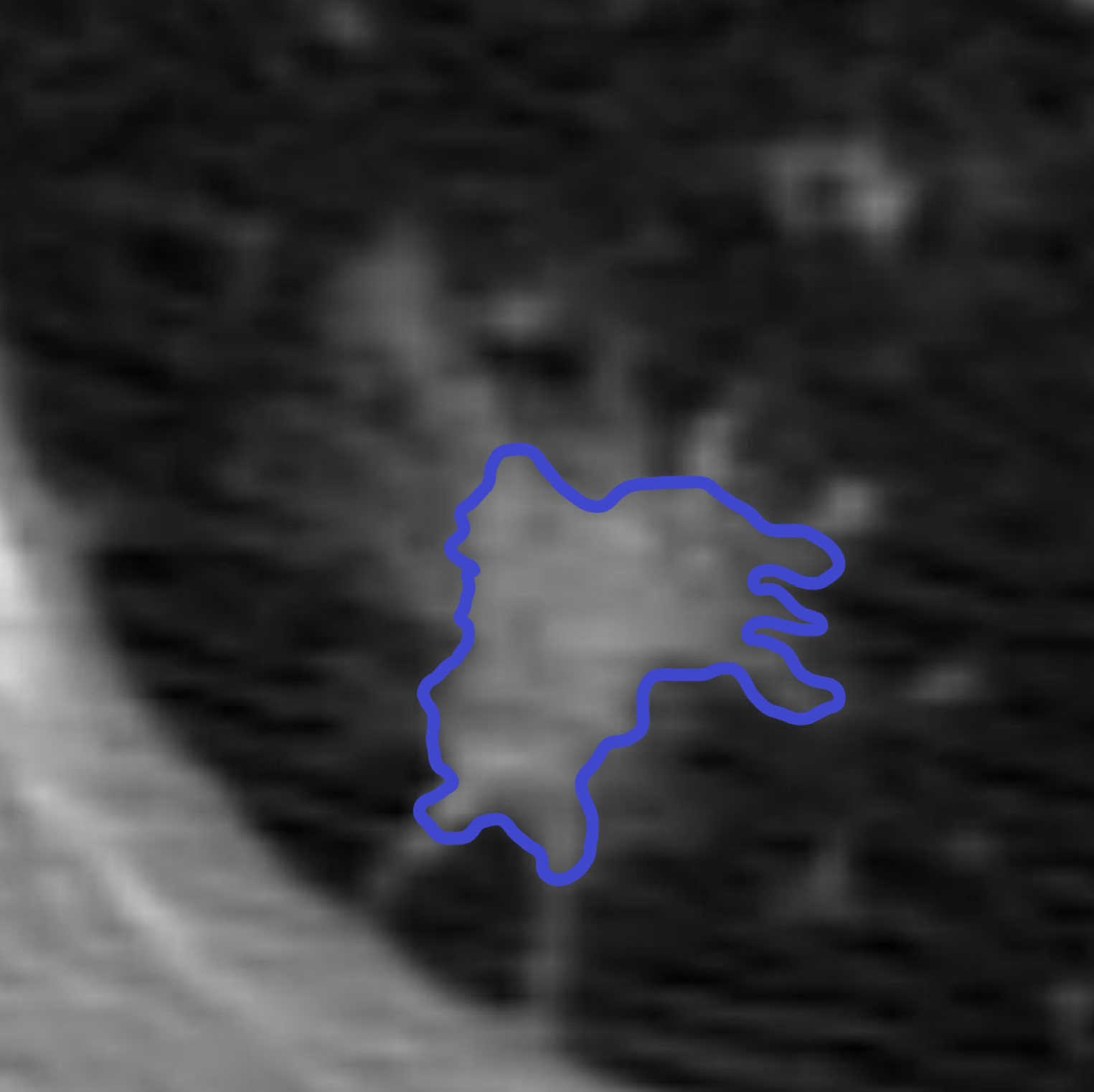} &
        \includegraphics[width=2.6cm, height=2.6cm]{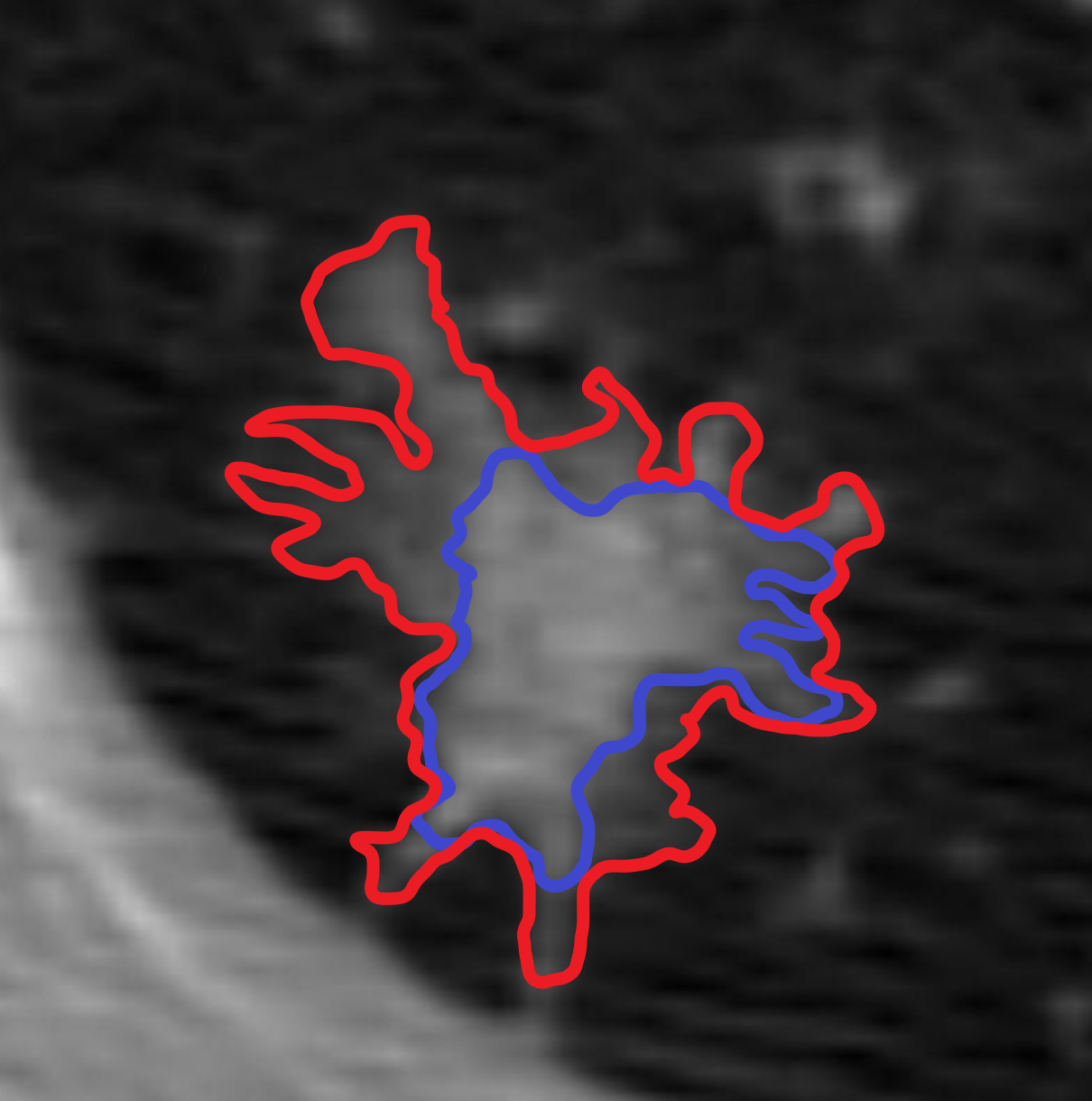} \\
        &\small\circled{1} Draw min.&\small \circled{2} Draw max.
    \end{tabular}
    \setlength\tabcolsep{6pt}
    \caption{The two steps of the process for producing Confidence Contours annotations, demonstrated on a sample from LIDC.}
    \label{fig:cc_steps}
\end{figure}


While such models provide spatial uncertainty distributions, they are not able to communicate the optimal thresholds for which medical decision-makers can make reliable and informed decisions. Humans have cognitive biases when interpreting uncertainty in the form of probability distributions~\cite{Tversky1974JudgmentUU}. Without the guidance of thresholds on these distributions, humans may make inaccurate inferences. Domain researchers who have attempted to apply uncertainty-aware models have reported similar challenges for reliable interpretation~\cite{Jungo2019AssessingRA,Ng2020EstimatingUI}.

Instead of a model-centric approach that utilizes complex model mechanisms to infer uncertainty from singular boundary annotations, we propose a data-centric approach~\cite{Hamid2022FromMT} in which models are trained on a novel annotation representation that directly communicates human-defined uncertainty labels.
In this paper, we present Confidence Contours (CCs), a novel semantic segmentation annotation representation that involves a pair of annotations---one forming a `contour' of high confidence (the `min') and another forming a `contour' of low confidence (the `max'). We also introduce a procedure for annotators to create CCs (Figure~\ref{fig:cc_steps}).


We conducted an annotation study to create segmentations on two datasets: the Lung Image Dataset Consortium (LIDC) with expert annotators (medical students), and a synthetic dataset, FoggyBlob, with non-expert annotators.
We evaluate four aspects of CCs, corresponding to four stages of the modeling pipeline from annotation to judgement-making.
First, we evaluate the usability and efficiency of our annotation system through a self-reported survey on task load and by measuring the time taken to complete annotation tasks. 
Second, we evaluate CCs' representative capacity by analyzing how well a CC annotation encompasses reference sets of singular annotations.
Third, we train almost 300 general segmentation models with varying architectures to robustly investigate how CC labels can be learned and predicted in modeling.
Last, we interview five medical experts to understand how CCs compare to alternative methods in terms of interpretability and utility for diagnosis.

From our annotation study, users reported an increase in self-reported task load while creating CCs; however, no task load metric saw more than a 1.3 point increase on the 10-point Likert scale used. 
Examining the logs of our annotation tool, we find that the time taken to produce a min \textit{and} max contour (44 sec for LIDC; 45 sec for FoggyBlob) was unsurprisingly more than the time taken to create a singular annotation (27 s for LIDC; 25 s for FoggyBlob), but not as much as double.
Moreover, annotators produced statistically significantly more consistent CCs than singular annotations.
Compared to a baseline constructed using singular annotations, CCs had a statistically significantly lower representation error ($-$19.9\% for LIDC and $-$41.8\% for FoggyBlob)---reflected through the ability to encompass a set of singular annotations.  
We show that general segmentation models are capable of learning to predict CC annotations to a similar degree of performance as standard singular annotations.
Finally, in our interviews, we find that medical experts generally prefer CC-type segmentations over singular masks and continuous uncertainty masks due to how they capture \textit{structural} uncertainty. 
In addition, they find CCs more interpretable and useful for judgement-making than continuous maps.
We conclude with discussions of the implications of our work for visual uncertainty modeling problems.


%
%



\section{Related Work}



\subsection{Medical Uncertainty Modeling}
\label{place:prev_work_ur}



Existing work to address uncertainty representation in segmentation is dominated by model-centric approaches, which retain singular annotation data (used in standard segmentation training) but develop novel modes of model interaction, design, and training. Elicitation-based approaches alter internal or external network states, such as through dropout~\cite{Gal2015DropoutAA,EatonRosen2018TowardsSD} or test-time augmentation~\cite{Wang2018AleatoricUE}, and composite the variation in predictions into an uncertainty map. Multi-candidate approaches~\cite{Rupprecht2016LearningIA,Ilg2018UncertaintyEF}
develop specialized training procedures to allow models to predict multiple hypotheses. Generative approaches, on the other hand, use probabilistic sampling and allow for the production of a theoretically infinite quantity of candidate predictions~\cite{Kohl2018APU,Baumgartner2019PHiSegCU,Monteiro2020StochasticSN}.

Surveys of uncertainty modeling in medical segmentation find that lack of contextualization and interpretability pose serious problems for the application of these approaches in practice. While uncertainty predictions perform well in dataset-wide metrics, they may not be coherent on a per-subject basis~\cite{Jungo2019AssessingRA}. Pixel/voxel-wise uncertainty measures produced by such models are biased towards producing `smooth' uncertainties within local regions, which lead to non-negligible errors~\cite{Jungo2020AnalyzingTQ}. Moreover, it is difficult to ascertain the uncertainty over complete structures rather than over voxels, which can have high variation within structures~\cite{Vasiliuk2022ExploringSU}. It has also been shown that per-voxel uncertainty measures in elicitation approaches can be highly dependent on modeling parameters rather than inherent uncertainty, and therefore pose challenges for clear interpretation~\cite{10.1007/978-3-031-16749-2_3}.

At the same time, researchers in medicine emphasize that machine learning applications cannot be purely computational and need to be designed to provide interpretations in addition to predictions~\cite{Chen2021ExplainableMI}. To foster effective human-AI collaboration, such interpretations need to take into account and support medical decision-makers' cognitive processes~\cite{Rundo2020RecentAO,Antoniadi2021CurrentCA}. 
To develop models that learn to predict more concrete and transparent signals,
recent work has proposed using cross-annotator disagreement as a directly-given measure of aleatoric uncertainty, which does not require models to implicitly infer distributions~\cite{Hu2019SupervisedUQ,Fornaciari2021BeyondB}. We take inspiration from such work in our data-centric approach.

\subsection{Capturing Uncertainty in Annotation}

On the annotation side, researchers have explored new designs for annotation systems and workflows that can capture uncertainty during the annotation process and improve consistency. 
Prior research has found traditional single-label annotation to be insufficient for identifying uncertainty in annotations, with proposed improvements including simple adjustments such as allowing multiple answers from each annotator~\cite{Jurgens2013EmbracingAA}, or asking for self reported confidence \textit{distributions} over the set of answers~\cite{Chung2019EfficientEA,Collins2022ElicitingAL}.
Others instead view the concept of fixed answer choices to be itself deficient. Some researchers propose the use of rationales as answers~\cite{donahue-iccv11,McDonnell2016WhyIT}, while others  propose open-ended answers that are then clustered or taxonomized~\cite{Kairam2016PartingCC,chang2017revolt}.
Finally, some have proposed more middle ground solutions in the form of new annotation representations, such as \textit{ranges} in scalar rating annotation~\cite{Chen2021GoldilocksCC}. Instead of asking for confidence or uncertainty via a question separate from the annotation itself, range annotations enable annotators to directly convey uncertainty calibrated to their annotation. This approach requires relatively low effort while also improving consistency in annotations. 
With Confidence Contours, we engage with uncertainty through a similar lens, where annotators are directly conveying uncertainty by providing a ``range'' annotation in two dimensions over an image.

\section{Proposed Approach}
\subsection{Confidence Contours}
In the standard segmentation annotation paradigm, singular masks for model training are derived by aggregating annotations made by multiple annotators. On the other hand, to produce Confidence Contour (CC) annotations, a single annotator provides two annotation `contours': a `min' contour and a `max' contour. 
The min contour is the set of pixels in the positive class with high confidence. The max contour is the set of pixels in the positive class with at least low confidence, and therefore spatially encompasses the min contour. Intuitively, the min represents what is `definitely' in the positive class, and the max what is `possibly'.
It follows that the region `outside' of the max ($\neg$ max) represents the set of pixels in the negative class with high confidence. 
The region spatially `between' the min and the max contours specifies the range of theoretically plausible singular annotations, and can be conceptualized as spatial bounds on the `true' distribution of all singular annotations. 
In cases where there is no ambiguity, the min and the max contours are equivalent and behave as singular annotations; therefore, singular annotations can be conceptualized as a subset of CC annotations. We formalize this below in ``Evaluating Representative Capacity.''
We use CCs on a binary segmentation problem in this paper for simplicity, in which each pixel belongs to either a positive or negative class, although it can be trivially adapted for multi-class segmentation problems.

\begin{figure}
    \centering
    \setlength\tabcolsep{1.5pt}
    \begin{tabular}{ccc}
        \includegraphics[width=2.6cm]{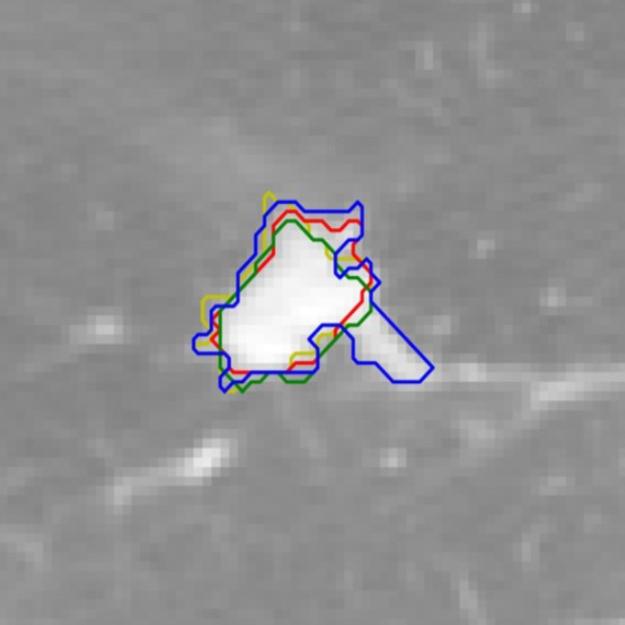} & 
        \includegraphics[width=2.6cm]{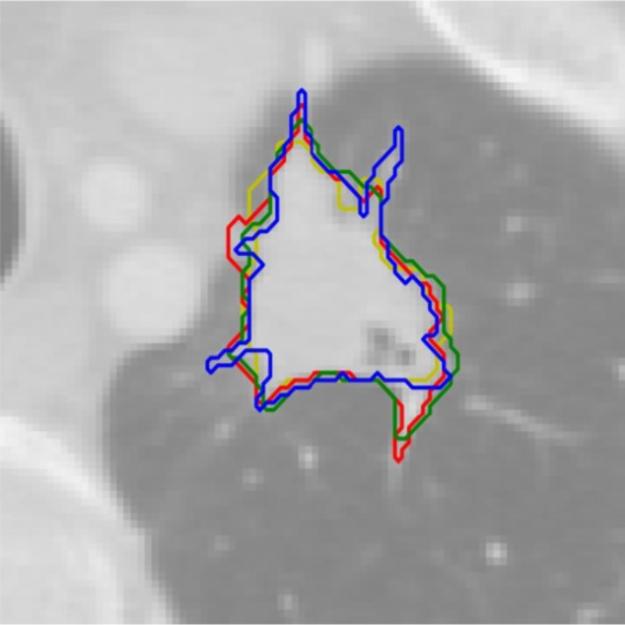} &
        \includegraphics[width=2.6cm]{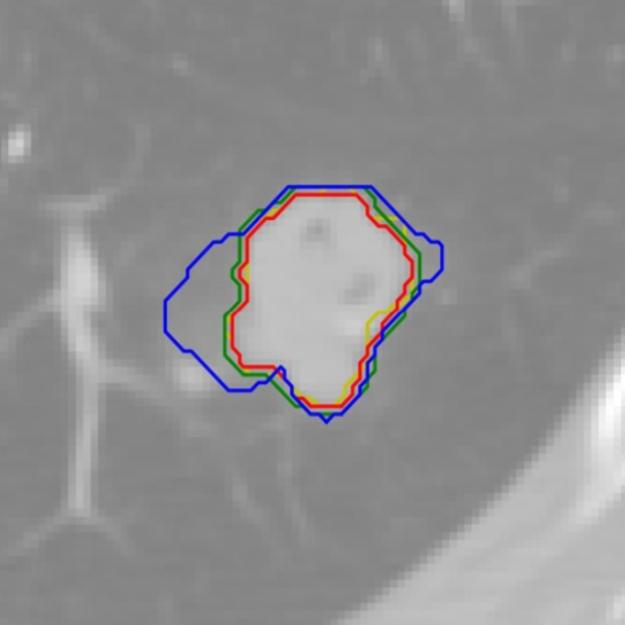} \\
        \includegraphics[width=2.6cm]{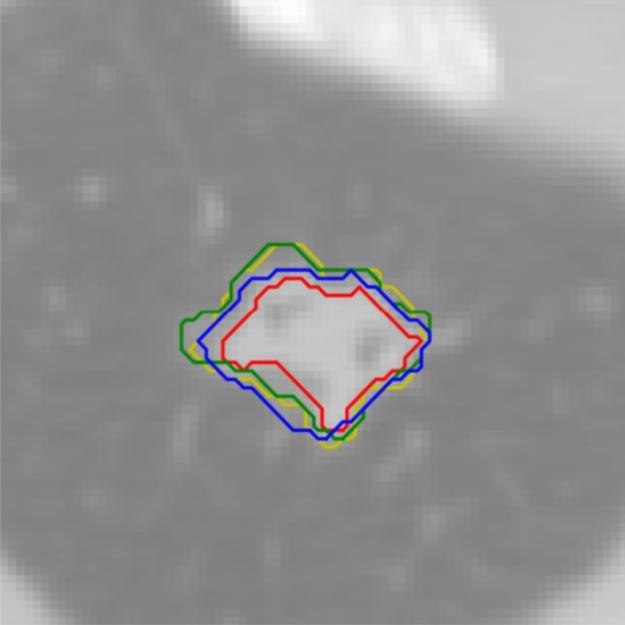} &
        \includegraphics[width=2.6cm]{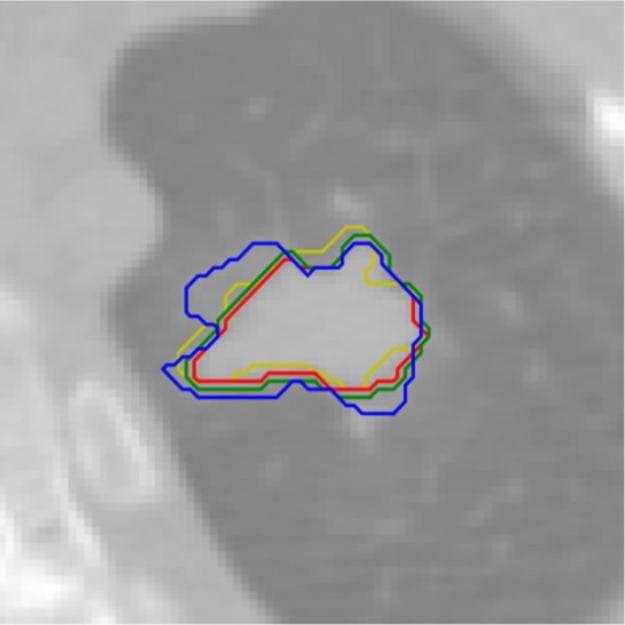} &
        \includegraphics[width=2.6cm]{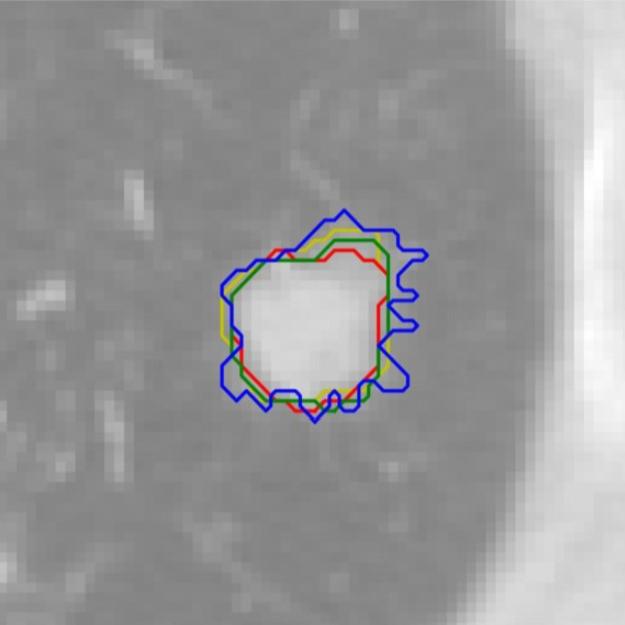}
    \end{tabular}
    \setlength\tabcolsep{6pt}
    \caption{Several sample high-disagreement images from LIDC. Disagreement follows a ``min/max'' structure: while most annotations cover a shared region, some annotations extend in concentrated regions. Each colored annotation represents a singular boundary annotation made by a different annotator.}
    \label{fig:lidc_disagreement}
\end{figure}

\subsection{Annotation Process}

Our annotation interface extends that of traditional singular annotation.
Annotators are given a poly-line tool that they use to draw polygon bounds indicating a segmentation.
Regions drawn by the poly-line tool can be edited by ``adding'' or ``subtracting'' from the current segment region. 

To produce CC annotations, annotators first draw a min contour. Annotators can adjust the min contour region until they are satisfied. Next, annotators press a button to make a copy of the min contour as the initial state of the max contour. Annotators then progressively add regions of low confidence to enlarge the max contour (Figure~\ref{fig:cc_steps}). Restated, the max contour is defined in terms of spatial additions to the min contour. This ordering of steps in this process reflects the nature of disagreement in many medical segmentation tasks: disagreement is concentrated in `controversial' regions concerning the inclusion or exclusion of particular ambiguous structures (Figure~\ref{fig:lidc_disagreement}).



\section{Evaluations}


In order to evaluate Confidence Contours throughout the modeling pipeline from annotation to judgement-making, we break down our evaluation into four components.
\begin{enumerate}
    \item \textbf{Annotation Study:} Is the annotation tool and two-step annotation process is usable and efficient for annotators? Are annotators able to produce more consistent annotations? We conduct a within-subjects annotation experiment with 45 participants who annotated 600 images using both the CC and singular method.
    \item \textbf{Representative Capacity Analysis:} How much uncertainty information do CCs represent, compared to singular annotations? We define several metrics and compare across collected CC and singular annotations.
    \item \textbf{Modeling Viability Analysis:} Do segmentation models trained on CCs produce viable masks? Is there loss in performance, compared to training on singular annotations? We train 144 segmentation architectures on both CCs and singular annotations, systematically comparing performance across different model sizes and types.
    \item \textbf{Interpretability and Utility for Medical Experts:} Do medical experts prefer CCs, singular, or continuous segmentation maps, and for what reasons? Which maps do experts find to be more interpretable model explanations and more actionable for diagnosis? We recruited 5 medical experts to examine multiple lung nodule segmentations and interviewed them to compare the utility of different uncertainty representation methods.
\end{enumerate}



\subsubsection{Datasets}
\label{sec:datasets}

\begin{figure}
    \centering
    \setlength\tabcolsep{1.5pt}
    \begin{tabular}{ccc}
        \includegraphics[width=2.6cm]{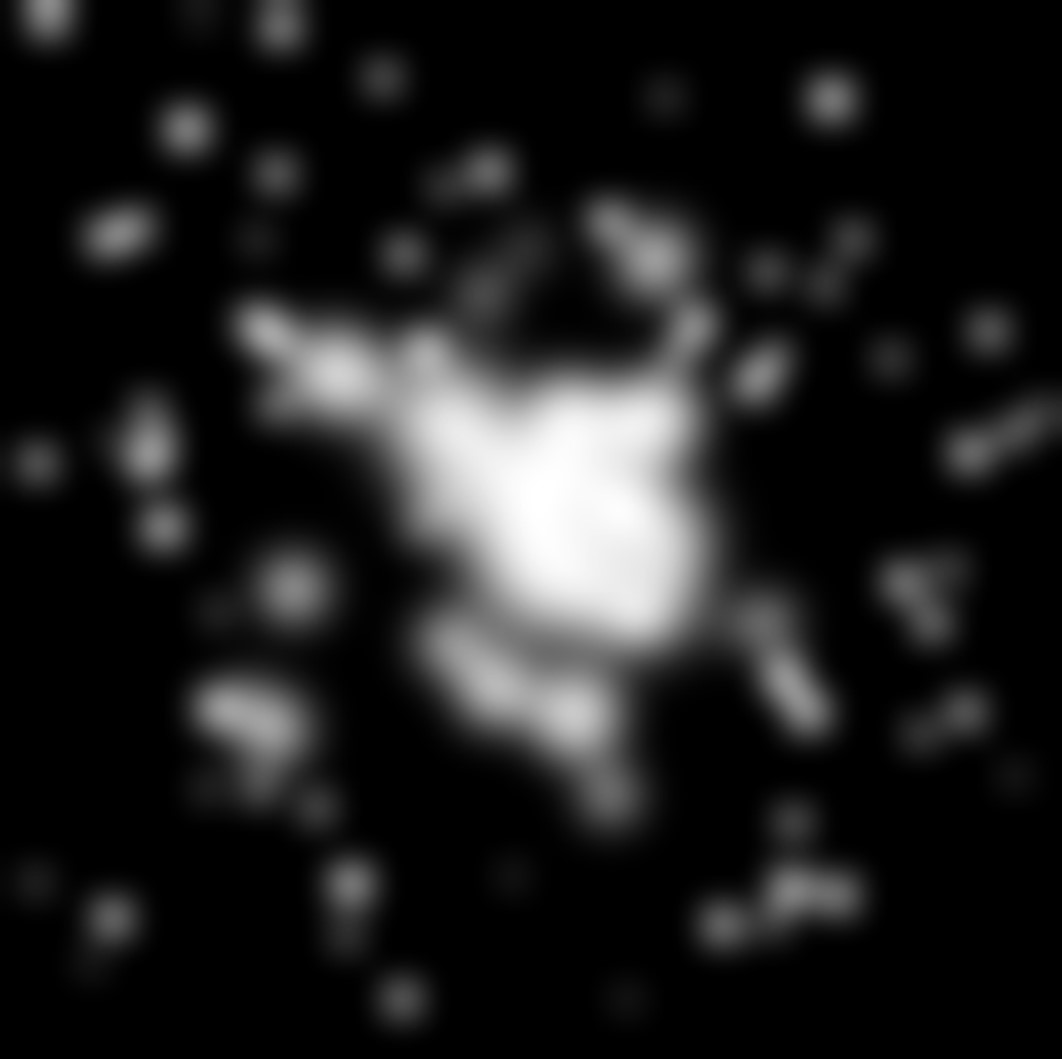} & 
        \includegraphics[width=2.6cm]{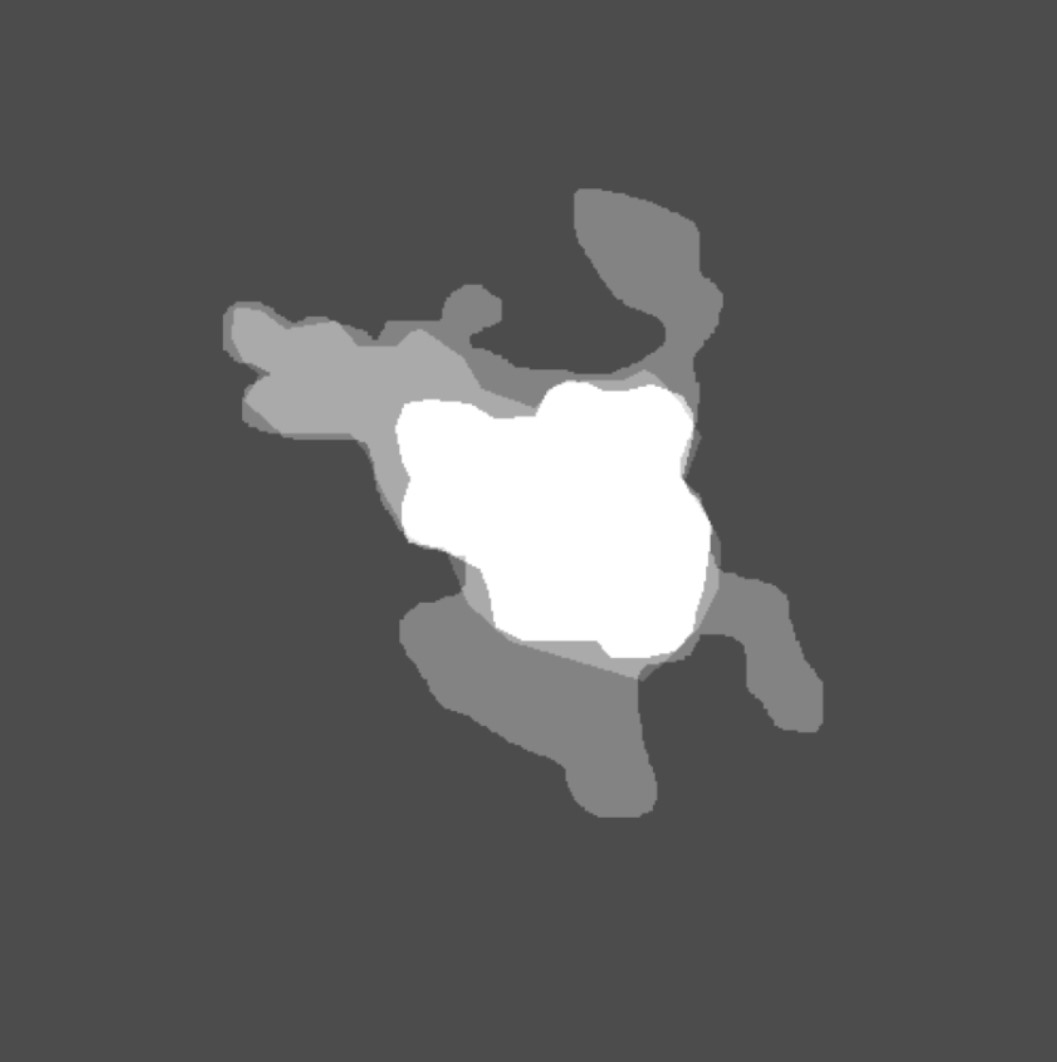} & 
         \includegraphics[width=2.6cm]{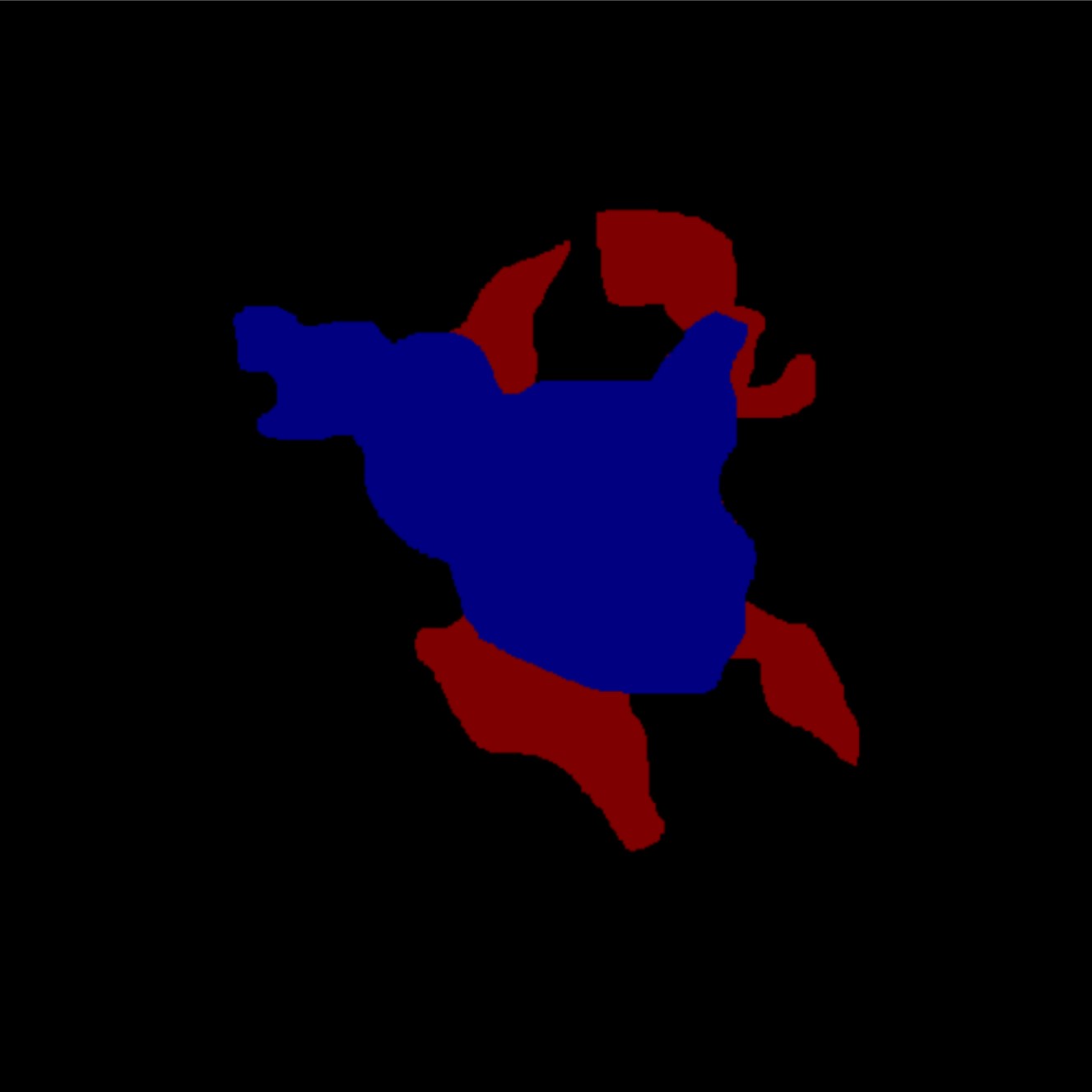} \\
         \includegraphics[width=2.6cm]{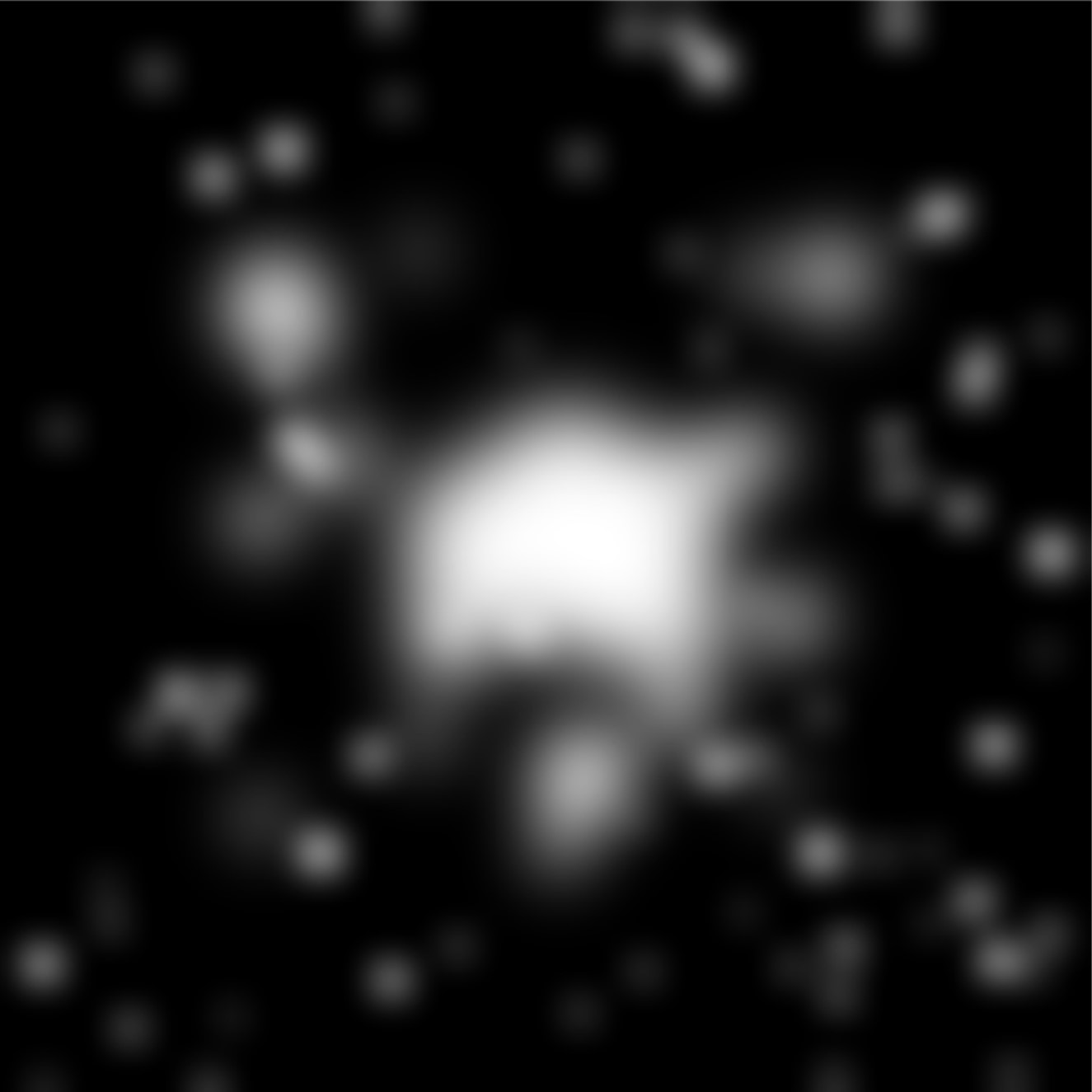} & 
         \includegraphics[width=2.6cm]{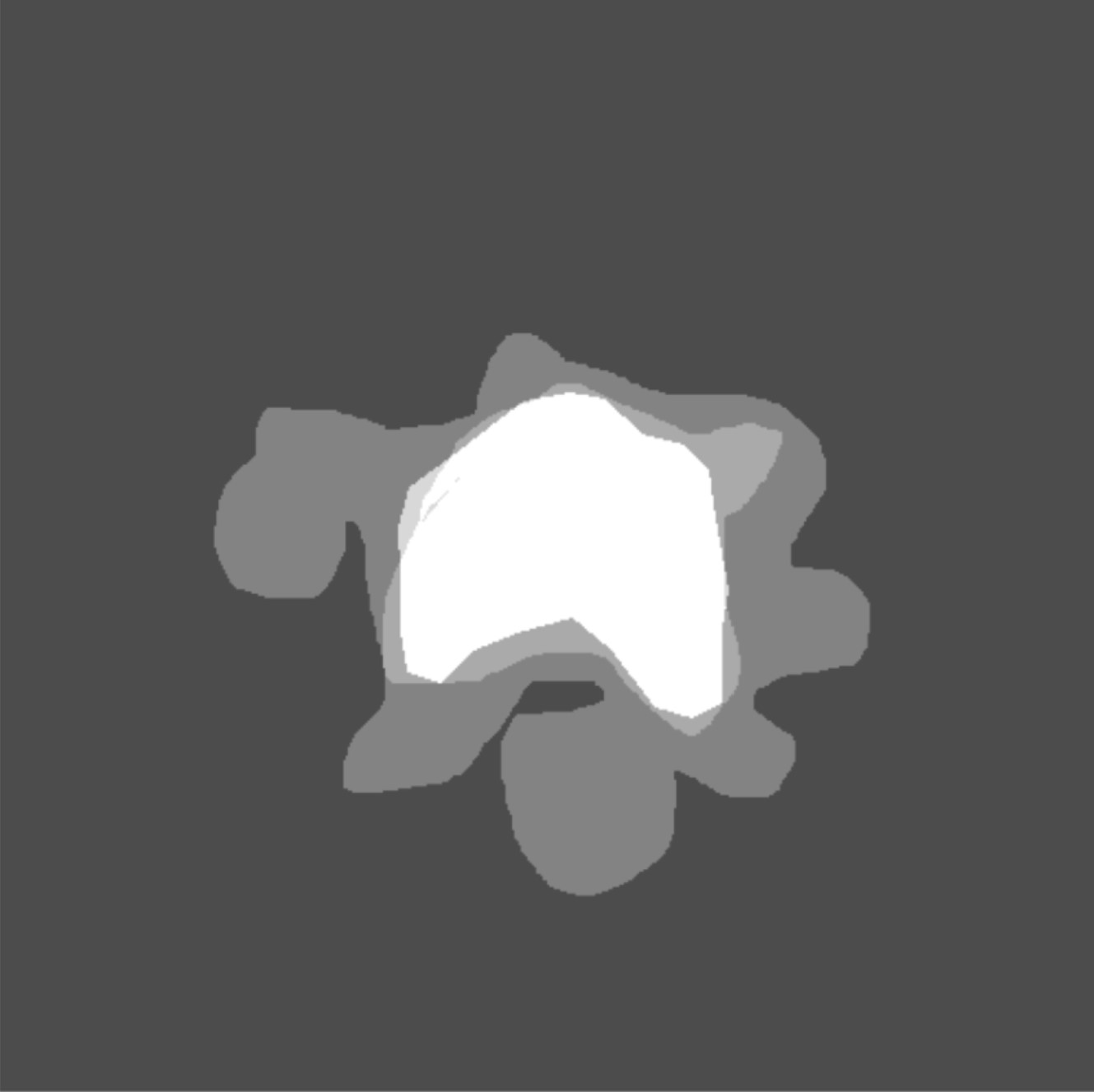} & 
         \includegraphics[width=2.6cm]{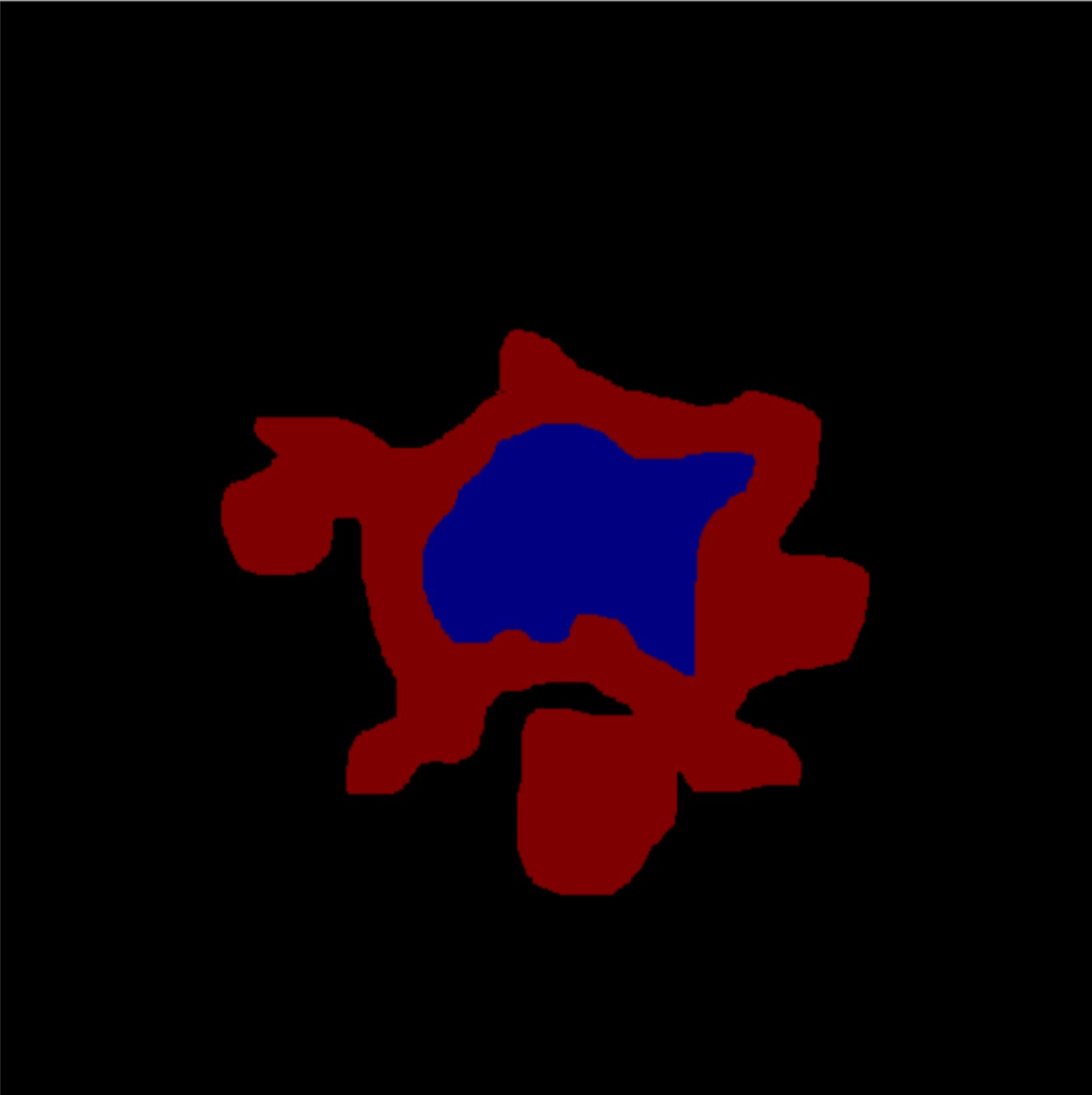}
    \end{tabular}
    \setlength\tabcolsep{6pt}
    \caption{Samples from the FoggyBlob dataset (left); composited maps of singular annotations obtained for that image (center); and CC annotations (right). Individual annotators' CC annotations generally reflect the distribution of disagreement in the composited singular annotation maps.
    }
    \label{fig:foggyblob}
\end{figure}

We assembled two datasets of images to conduct our evaluation experiments of Confidence Contours (CCs): a subsample from the Lung Image Database Consortium (LIDC) and FoggyBlob, a custom synthetic dataset.

LIDC contains clinical thoracic CT scans of pulmonary nodules with 2--4 annotations from professional radiologists for each image. Irregularities in the size and shape of pulmonary nodes can be a strong indicator for lung cancer and other conditions
\cite{Loverdos2019LungNA}.
This makes LIDC an effective demonstrative case of an ambiguous and high-stakes problem, since cumulative disagreements between annotators over the inclusion or exclusion of particular crucial structures can substantively influence the diagnosis. We only consider `windows' in which a pulmonary node is guaranteed to be present to focus strictly on segmentation rather than localization or detection, which we consider to be a separate problem. This separation is common in medical imaging diagnosis problems where there is a large discrepancy in scale between the image size and the annotated object(s)~\cite{Li2022DeepLA}.
For our experiments, we sampled the 400 highest-disagreement windows in LIDC, with disagreement measured as the intersection over union (IoU) between all dataset-provided annotation masks. 

The FoggyBlob dataset (Figure~\ref{fig:foggyblob}) is a synthetic dataset we created to simulate the challenges of segmenting in ambiguous contexts for layperson annotators. Each image is composed of a prominent centered mass and multiple blurred `branches.' 
The annotator's task is to annotate the central mass, and therefore implicitly to decide whether or not certain `branches' are fit for inclusion. FoggyBlob replaces the technical knowledge required for making such inclusion/exclusion determinations in medical segmentation problems with human intuition for objectness.

\subsection{Evaluating the Annotation Process}

\subsubsection{Method}

\begin{table}[t]
    \centering
    \begin{tabular}{l|rr|rr}
        \toprule
        Dimension &
          \multicolumn{2}{c}{LIDC} &
          \multicolumn{2}{c}{FoggyBlob} \\
        & Singular & CC & Singular & CC \\
        \midrule
        Mental Demand & 3.7 & *4.9 & 3.3 & *4.6 \\
        Physical Demand & 2.7 & 3.3 & 3.9 & 3.7 \\
        Temporal Demand & 4.2 & *4.9 & 5.0 & 5.5 \\
        Performance & 6.9 & 6.9 & 6.8 & 6.9 \\
        Effort & 4.8 & *5.7 & 5.0 & 5.1 \\
        Frustration & 3.0 & *4.2 & 2.7 & *4.0 \\
        \bottomrule
    \end{tabular}
    \caption{Average annotator responses across six dimensions and two datasets on the experience annotating using singular and CC methods, evaluated on a 10 point scale (1$=$``very low", 10$=$``very high"). * indicates a statistically significant relationship, measured with a relative $t$-test by annotator.}
    \label{tab:nasatlxtable}
\end{table}

We recruited 30 undergraduate students majoring in the biological sciences to each annotate 40 images with CCs and 40 images with the singular method as a baseline. 
Every image was annotated by 3 different annotators, producing 3 CC and 3 singular annotations. Annotators were briefed and evaluated on the relevant radiological and medical background before annotating. 
We received IRB approval for this study under ID 16518.
While our annotators were students, we consider them expert annotators as the quality of our annotations generally matched that of the LIDC dataset produced by radiologists.
There was no statistically significant difference between the level of disagreement comparing our annotations and LIDC annotations versus comparing LIDC annotations with each other.
This similarity in skill level may be explained by our task only being the simpler \textit{segmentation} rather than the more challenging \textit{localization} task.

The same study procedure was used for the FoggyBlob dataset with 15 students from general areas of study. 


To counteract bias from learning effects, we counterbalanced the order of annotation method (singular and Confidence Contours) for annotation studies on both datasets.
All participants were compensated $\$17$ per hour, higher than the local minimum wage at the time of study. Of 45 total annotators, 69\% identify as female and 31\% identify as male.
In total, we collected 3,600 annotations across 600 images.\footnote{See https://github.com/andre-ye/confcontours-data for annotation dataset.}

After annotating the assigned image set for each annotation method, annotators completed a NASA TLX questionnaire on task load~\cite{Hart1988DevelopmentON}. 
The survey recorded the following dimensions on a 10-point Likert scale: mental demand, physical demand, temporal demand, performance, effort, and frustration. We also measured the time annotators spent per annotation through logging in our annotation tool.


\subsubsection{Annotating CCs is not considerably more burdening than singular ones.}
Annotators generally report that producing singular annotations requires lower overall load than producing CCs (Table~\ref{tab:nasatlxtable}). This is expected because creating CCs requires physically more user input than the singular method. Notably, however, annotators did not experience statistically significant differences in self-perceived performance level across both tasks ($p < 0.05$), suggesting that CC annotation is learnable. Moreover, there are no statistically significant differences between the two annotation methods except in the mental demand and frustration dimensions for FoggyBlob. This suggests that the significance of task-load differences may be dependent on the complexity of the context. Last, the absolute magnitude of the differences between singular and CC annotations are at most 1.3 points out of a 10-point scale in all dimensions.


\subsubsection{CCs are efficient to annotate relative to the amount of information produced.}
We measure the average time taken to produce CCs versus singular annotations after the first 20 annotations, to account for the period where the annotator is learning the tool and task.
On average, annotators spent 27 sec to create a singular annotation and 44 sec to create a CC annotation per image in LIDC. For FoggyBlob, the time spent was 25 sec and 45 sec, respectively.
Across all annotators, the annotation time for each sample using CCs is 68.50\% and 75.32\% more than using the singular method for the LIDC and FoggyBlob datasets, respectively. 
Thus, having one annotator create a CC annotation is faster than having at least two annotators each create a singular annotation and then use their disagreement to infer uncertainty.

\subsubsection{Annotators produce CCs more consistently than singular annotations.}
\begin{table}
    \centering
    \begin{tabular}{l|r|r|r}
        \toprule
        Dataset & Singular & Min & Max \\
        \midrule
        LIDC & 0.7296 & *0.6035 & 0.7301 \\
        FoggyBlob & 0.6261 & *0.5485 & *0.5555 \\
        \bottomrule
        
    \end{tabular}
    \caption{Disagreement, measured as the mean pairwise discrete Frech\'{e}t distance in pixel space, scaled by the mean longest chord in the annotation for approximate bounding, within groups of singular annotations, min contours, and max contours. * indicates statistically significant ($p <0.05$) decrease compared to the singular annotations' disagreement using relative $t$-test.}
    \label{tab:disagreementtable}
\end{table}

As another dimension of representative capacity, we evaluate how consistently annotators create the upper and lower bounds. Given that annotator disagreement has been shown to have strong relationships with sources of structural visual uncertainty, we use disagreement as a proxy for CCs' ability to represent uncertainty. 
We compute the disagreement between a set of annotations as the average pairwise discrete Frech\'{e}t distance, a common measure of curve similarity~\cite{Wylie2013THEDF}. 
In choosing a particular measurement, we are primarily concerned with the annotator behavior on the inclusion or exclusion of particular ambiguous structures along the surface of the curves. Area-based IoU measurements inflate the role of unambiguous regions of high agreement, so we use Frech\'{e}t distance instead.
Across both datasets, we observe a statistically significant reduction in disagreement between min contours as opposed to singular annotations (Table~\ref{tab:disagreementtable}). For LIDC, we find no statistically significant difference between disagreement between max contours and singular annotations, whereas we do for FoggyBlob. This is possibly due to the synthetic nature of the FoggyBlob dataset. 
On the other hand, in real-world datasets, different annotators may disagree on whether an ambiguous structure should be included into the max contour.
In general, we find that annotators tend to agree more on regions of high confidence than regions of low confidence.

\subsection{Evaluating Representative Capacity}

\begin{table}
    \centering
    \begin{tabular}{c|rr|rr}
        \toprule
        Dataset &
          \multicolumn{2}{c}{$\mathbb{L}^+$} &
          \multicolumn{2}{c}{$\mathbb{L}^-$} \\
         & CC & Base & CC & Base \\
        \midrule 
        LIDC & *0.1416 & 0.1659 & *0.1209 & 0.1615 \\
        FoggyBlob & *0.0837 & 0.1249 & *0.0622 & 0.1259 \\
        \bottomrule
    \end{tabular}
    \caption{Mean underflow and overflow across all samples in the LIDC and FoggyBlob datasets. *indicates statistically significant ($p<0.05$) compared to the baseline.}
    \label{tab:representative-capacity-results}
\end{table}

\subsubsection{Method}
One function of Confidence Contours (CCs) is as a means for one annotation to encompass the range of multiple singular annotation responses we might otherwise observe across a group of different annotators.
Existing work~\cite{10.1007/978-3-030-01364-6_12} shows that disagreements among annotators are often centered in regions of visual ambiguity, which cast uncertainty on the identification of relevant structures in the image. CC annotations should similarly be drawn to accommodate the same sources of uncertainty.
Additionally, in other annotation modalities, it has been shown that single annotators can often anticipate the distribution of responses of their peers~\cite{Chung2019EfficientEA}.

We can view any segmentation representation $s$ as a partition of an image into three types of points: $s^+$---points certainly associated with the subject-of-interest, $s^-$---points certainly \textbf{not} associated with the subject-of-interest, and $s^?$---points that \textbf{may or may not be} associated with the subject-of-interest.
Under this view, we can intuitively see that one segmentation $s_a$ \textit{bounds} another $s_b$ if $s_a^+ \subseteq s_b^+$ and $s_a^- \subseteq s_b^-$.
Of course, in practice, segmentation representations are rarely expected to perfectly bound another. To understand \textit{representative capacity}, we want to quantify the \textit{error} at which a segmentation fails to bound another.
To do this we define the following error metrics:
$$L^+(s_a, s_b) = |\{ \forall p : p \in s_a^+ \land p \notin s_b^+ \}|$$
$$L^-(s_a, s_b) = |\{ \forall p : p \in s_a^- \land p \notin s_b^- \}|$$
Intuitively, $L^+$ measures the degree to which $s_a^+$ fails to be a subset of $s_b^+$ (error in $s_a$ serving as a lower bound for $s_b$, or put simply, ``underflow'') and $L^-$ measures the degree to which $s_a^-$ fails to be a subset of $s_b^-$ (error in $s_a$ serving as an upper bound for $s_b$, ``overflow''). Together, they holistically represent error to which $s_a$ bounds $s_b$.

Of course, more practically, we would like to quantify the representative capacity of a new segmentation representation $s_a$ in relation to a reference \textit{set} of segmentations $S$. This can be done by computing the \textit{expected} error\footnote{We use $\neg s^-$ to denote $s^? \cup s^+$.}:
$$\mathbb{L}^+(s_a, S) = \mathbb{E}_{s \in S} [L^+(s_a, s) / |s_a^+ \cup s^+|]$$
$$\mathbb{L}^-(s_a, S) = \mathbb{E}_{s \in S} [L^-(s_a, s) / |\neg s_a^- \cup \neg s^-|]$$
Note that we introduce a normalization factor in the above expressions to account for the different sizes of $s^+$; this also bounds both metrics between 0 and 1 (inclusive).

We calculate the representative capacity of CCs ($\mathbb{L}^+_{\text{CC}}$ and $\mathbb{L}^-_{\text{CC}}$), using our singular annotations as a reference set.
As a baseline, we compute the representative capacity of singular annotations ($\mathbb{L}^+_{\text{base}}$ and $\mathbb{L}^-_{\text{base}}$) against the same reference set.
The data for evaluating singular annotations is constructed by taking a held-out annotation that was not used in the reference set.

\subsubsection{CCs exhibit lower underflow and overflow error.}

\begin{figure}
    \centering
    \setlength\tabcolsep{1.5pt}
    \begin{tabular}{cccc}
        \rotatebox{90}{LIDC}\hspace{3pt}
        &\includegraphics[width=2.6cm]{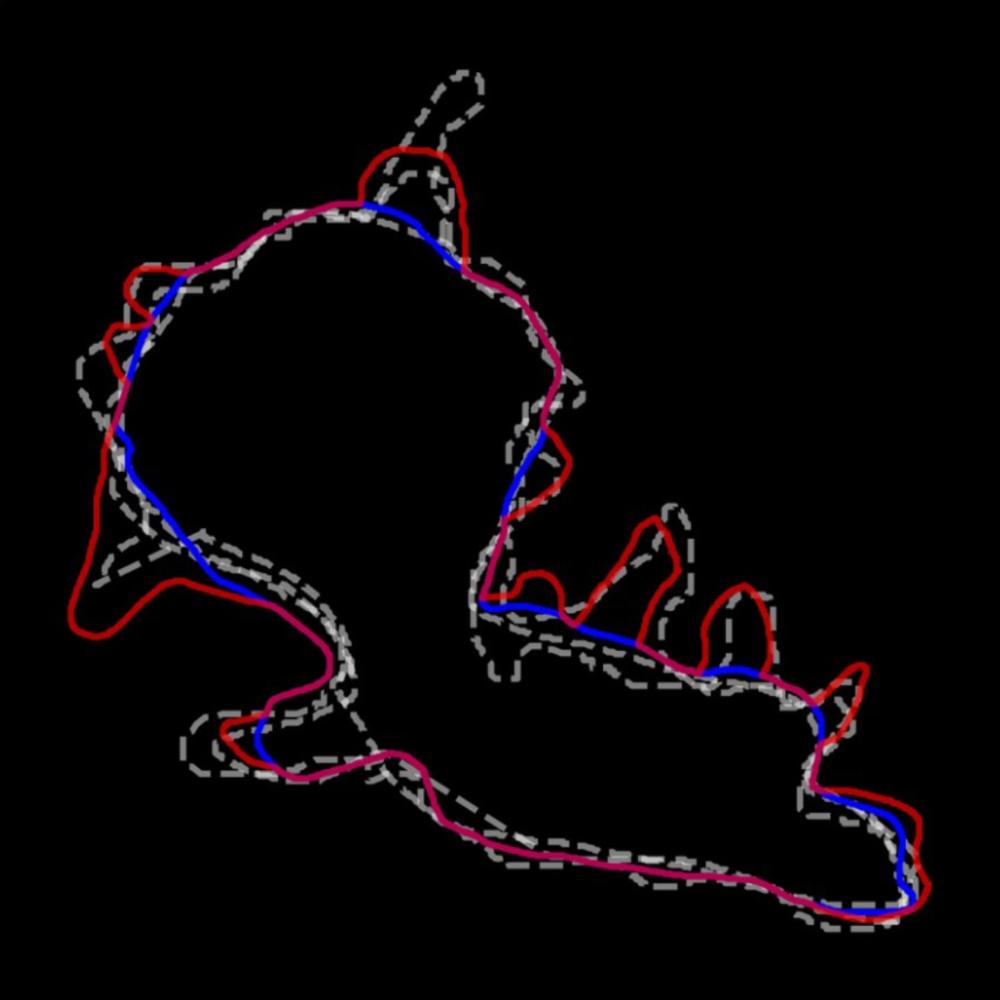}
        &\includegraphics[width=2.6cm]{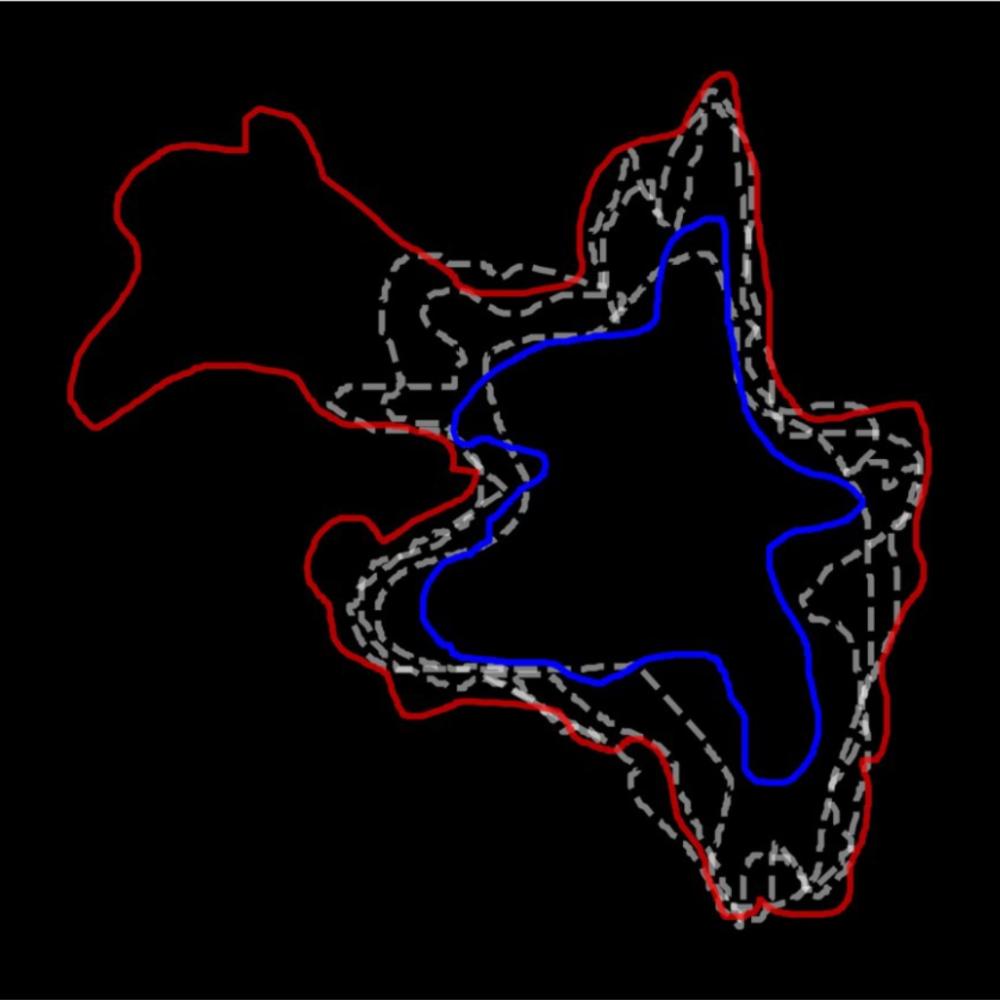}
        &\includegraphics[width=2.6cm]{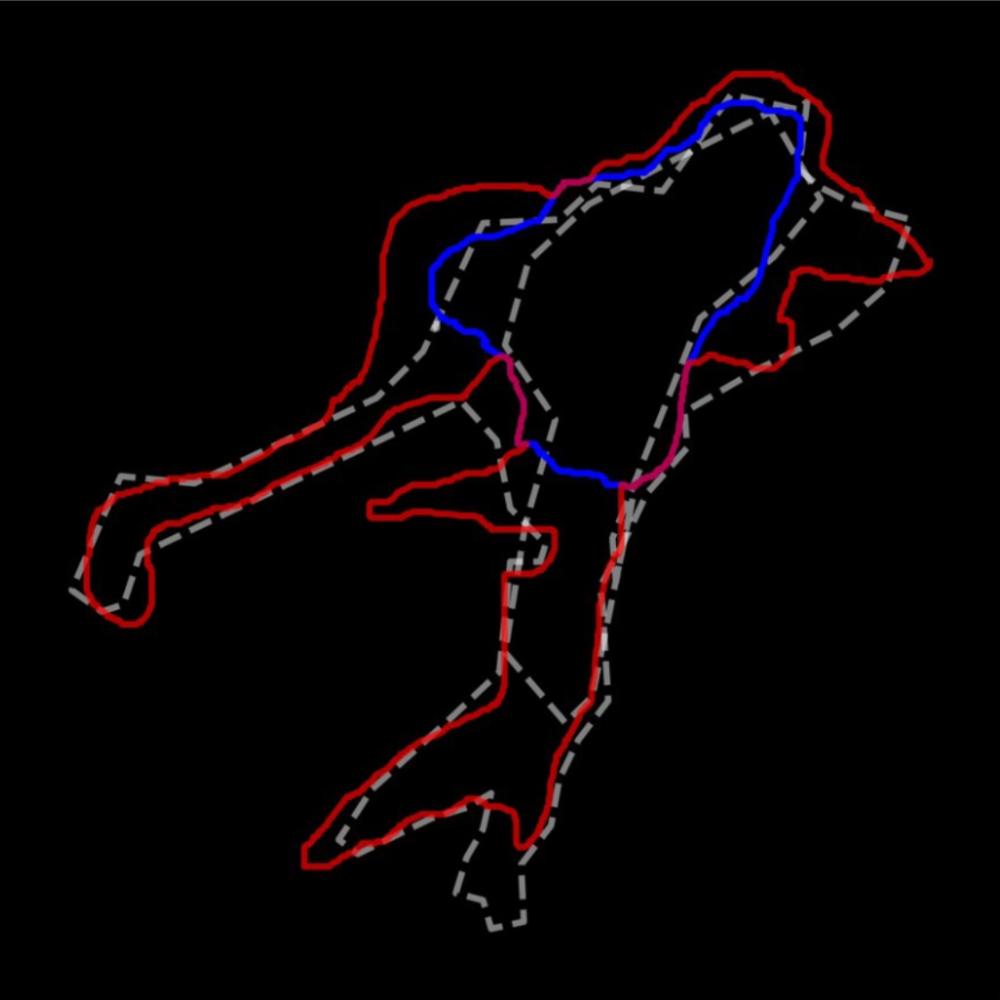} \\
         \small\rotatebox{90}{FoggyBlob}\hspace{3pt}&\includegraphics[width=2.6cm]{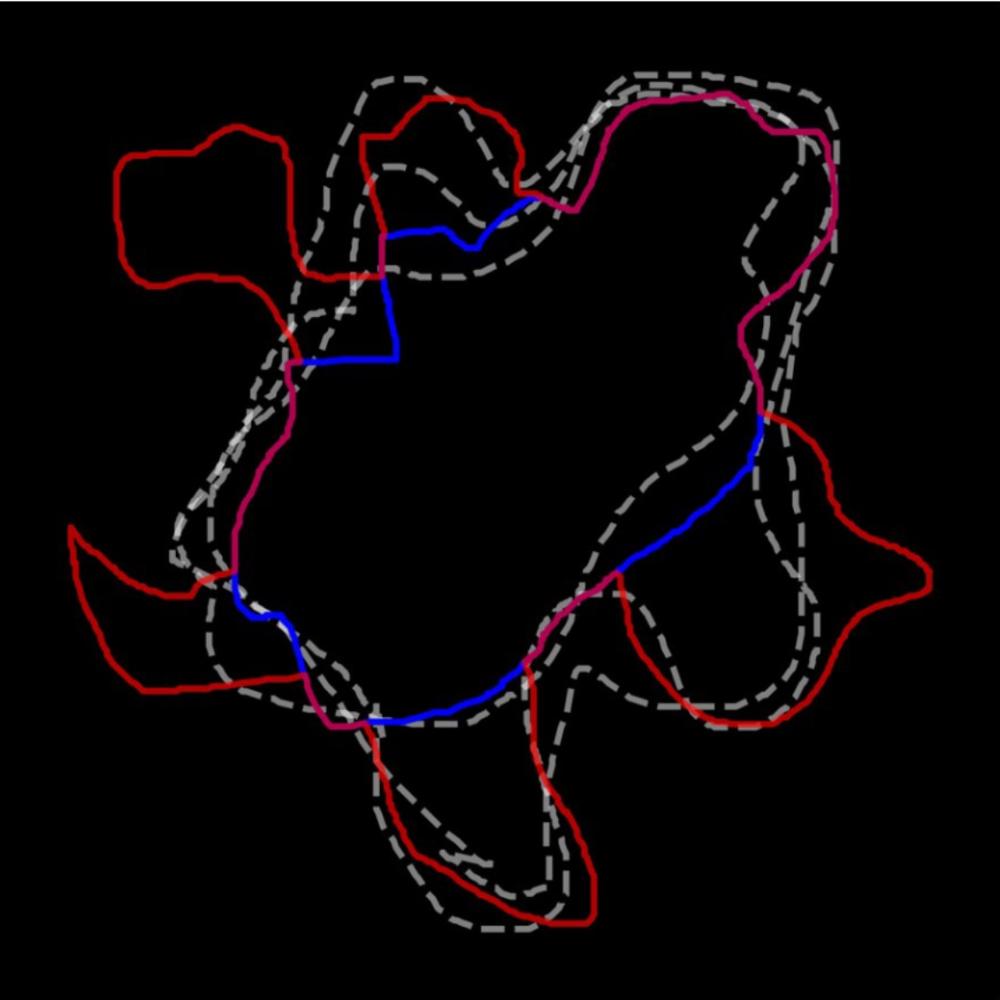} & 
        \includegraphics[width=2.6cm]{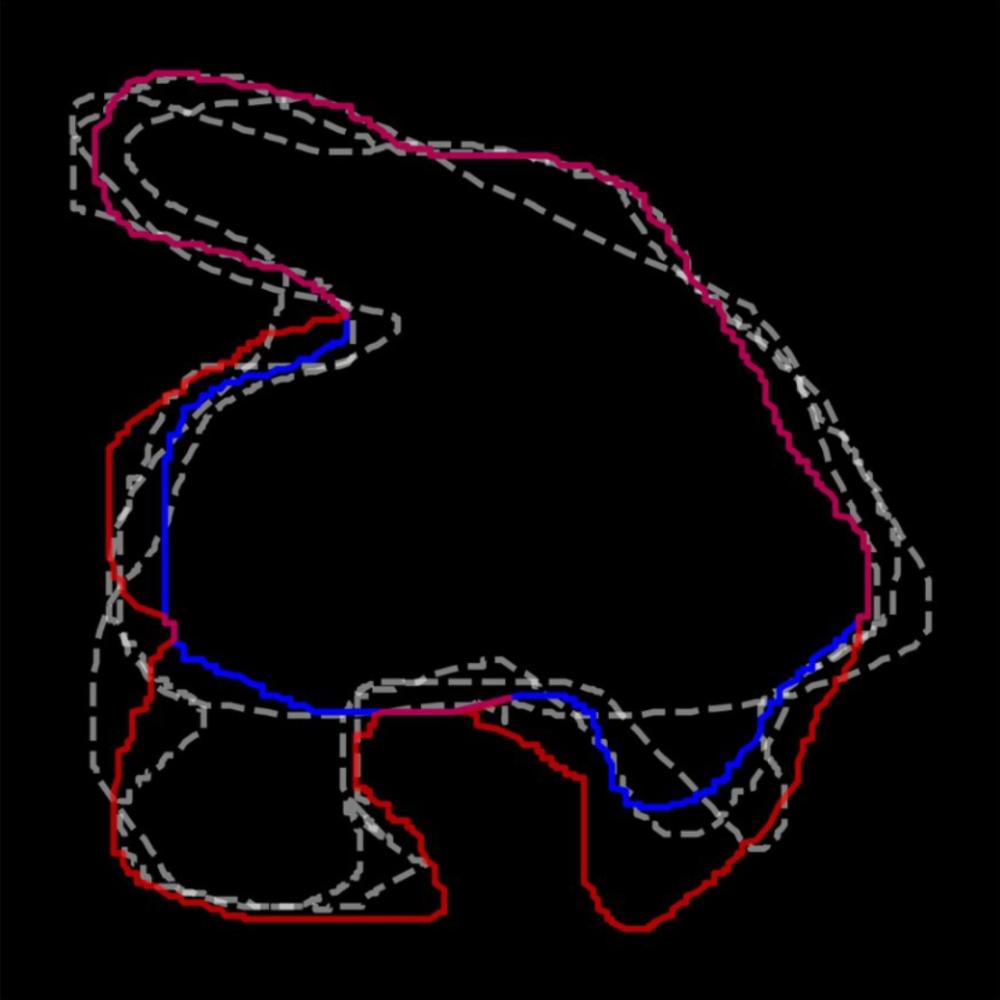} & 
         \includegraphics[width=2.6cm]{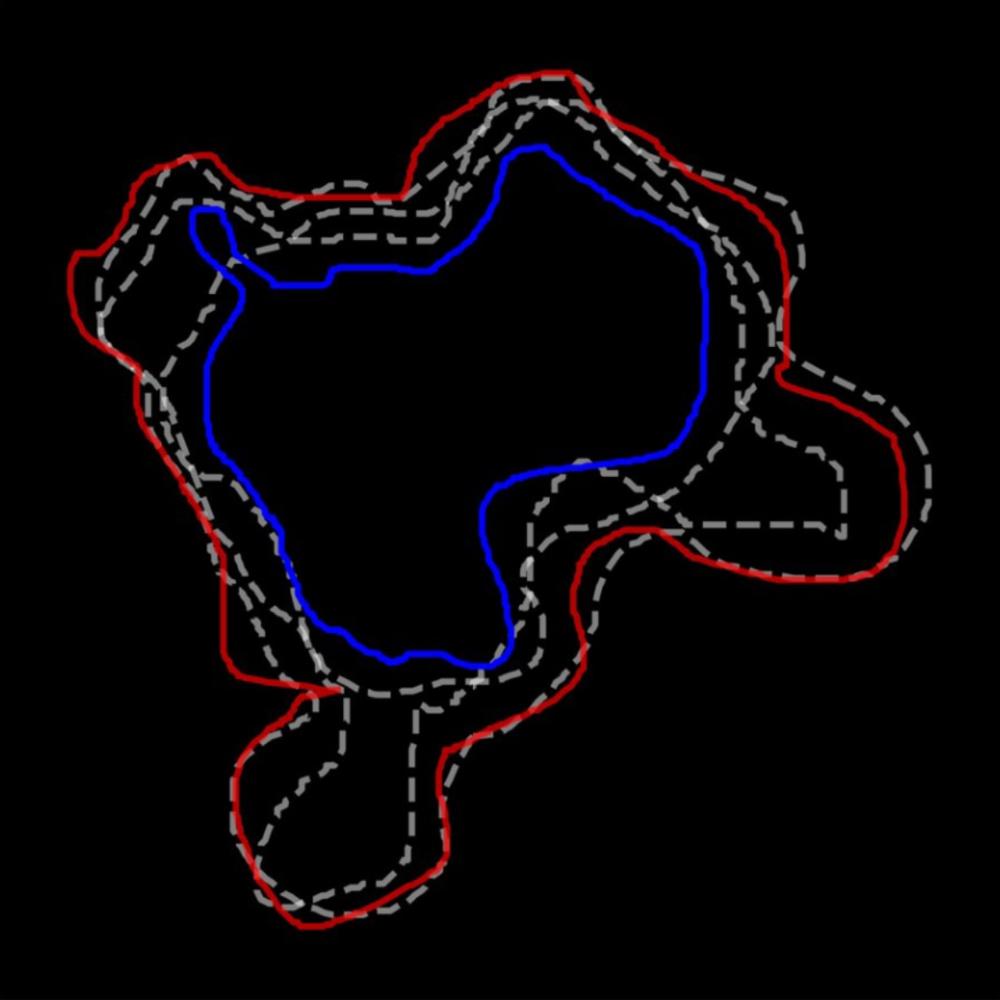}
    \end{tabular}
    \setlength\tabcolsep{6pt}
    \caption{Annotations from the LIDC (top) and FoggyBlob (bottom) datasets. The min (max) annotations are shown in blue (red), and ensembles of singular annotations are shown in dotted white.}
    \label{fig:representation}
\end{figure}

We find that, compared to the baseline, CCs demonstrate more representative capacity, as evidenced by statistically significantly lower underflow and overflow across both datasets (Table~\ref{tab:representative-capacity-results}). This representative capacity is clearly visualized in Figure~\ref{fig:representation}.
We also found that, for LIDC and FoggyBlob respectively, 42.96\% and 50.16\% of instances had $\mathbb{L}^+ \leq 0.05$ (a trivial level of error) and 40.20\% and 45.45\% of instances had $\mathbb{L}^- \leq 0.05$.
In addition to these observations, we found that for individual instances $s_a$ experience less overflow than underflow ($\mathbb{L}^-(s_a) \leq \mathbb{L}^+(s_a)$) to a statistically significant degree ($p=0.0236 < 0.05$ for LIDC, $p=0.0158 < 0.05$ for FoggyBlob) for LIDC and FoggyBlob respectively.

\subsubsection{The area of $s^?_{\text{CC}}$ correlates with disagreement across singular annotations.}
Moreover, we explored whether the degree of uncertainty in CCs correlates with the uncertainty observed from disagreements in sets of singular annotations. 
For each image, this is calculated as $|s^?_\text{CC}|$ a single CC annotation and $\mathbb{E}[|\neg s^-_\text{Base}|] - \mathbb{E}[|s^+_\text{Base}(x)|]$ for an ensemble of singular annotations.
We find that there is a statistically significant correlation between these metrics across both LIDC ($\rho=0.5868$, $p < 0.001$) and FoggyBlob ($\rho=0.4343$, $p < 0.001$).



These results suggest that $s^?_\text{CC}$ 
 represents bounds on the range of singular annotations; that is, we would expect a singular annotation drawn by some annotator to fall within $s^?_\text{CC}$. Overall, the results suggest that a single CC annotation represents many uncertainty-relating structural properties across multiple singular annotations.


\subsection{Evaluating Model Viability}

\subsubsection{Method}

We also explored whether CCs can be learned effectively by a variety of downstream segmentation models.
Conceptually, these models should predict both the min and the max contour masks simultaneously.
We simply train models to predict a two-channel mask for convenience, although there are several alternative methods to do so, such as training two separate models.
Therefore, any general segmentation model can be trivially modified to support CC labels. 
The optimization objective for a model $M$ using loss function $\mathcal{L}$ on CC-annotated data becomes:
$$\min_{M} [ \mathcal{L}(M_\text{min}(x), y_\text{min}) + \mathcal{L}(M_\text{max}(x), y_\text{max})]$$

We evaluated performance across four model architectures commonly used for medical segmentation: U-Net~\cite{Ronneberger2015UNetCN}, attention U-Net~\cite{Oktay2018AttentionUL}, DeepLab~\cite{Chen2016DeepLabSI} with a MobileNetV2 backbone~\cite{Sandler2018MobileNetV2IR}, and PSPNet~\cite{Zhao2016PyramidSP}.
To account for variability in training outcomes, we perform grid search over hyperparameters for each architecture: for U-Net and attention U-Net we tested configurations of batch size $\in$ \{8, 16, 32, 64\}, initial filters $\in$ \{8, 16, 32\}, and encoder-decoder pathway block count $\in$ \{2, 3, 4, 5\}; for DeepLab -- batch size $\in$ \{8, 16, 32, 64\}, filters $\in$ \{8, 16, 32\}; for PSPNet -- batch size $\in$ \{8, 16, 32, 64\}, initial filters $\in$ \{8, 16, 32\}, and block count $\in$ \{1, 2, 3\}. 
Model architectures were scaled down to accommodate for our smaller dataset and image size. For each architecture and hyperparameter set, we train one instance on singular annotations and another on CC annotations until convergence. We use the Adam optimizer and the dice loss optimization objective across all instances. 
Following standard practice, reference masks are generated from singular annotations using 50\% (majority) consensus; CC annotations are not aggregated (i.e., each image is associated with multiple labels). We employ an extensive augmentation pipeline, including affine transformations, blurring, and sharpening.


\subsubsection{Results}
\begin{figure*}[!t]
    \centering
    \setlength\tabcolsep{1.9pt}
    \begin{tabular}{c ccccc ccccc c}
        \small\rotatebox{90}{\parbox[c][0.2cm][c]{1.35cm}{Image}} \hspace{3pt}
        \includegraphics[width=1.35cm]{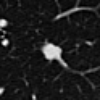} &
        \includegraphics[width=1.35cm]{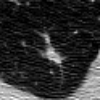} &
        \includegraphics[width=1.35cm]{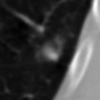} &
        \includegraphics[width=1.35cm]{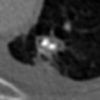} &
        \includegraphics[width=1.35cm]{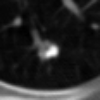} &
        \includegraphics[width=1.35cm]{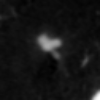} &
        \includegraphics[width=1.35cm]{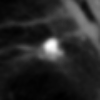} &
        \includegraphics[width=1.35cm]{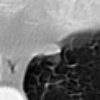} &
        \includegraphics[width=1.35cm]{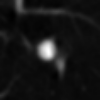} &
        \includegraphics[width=1.35cm]{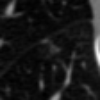} &
        \includegraphics[width=1.35cm]{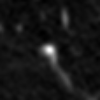} \\
        \small\rotatebox{90}{\parbox[c][0.2cm][c]{1.35cm}{Ground}} \hspace{3pt}
        \includegraphics[width=1.35cm]{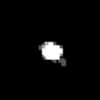} &
        \includegraphics[width=1.35cm]{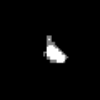} &
        \includegraphics[width=1.35cm]{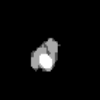} &
        \includegraphics[width=1.35cm]{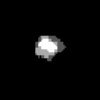} &
        \includegraphics[width=1.35cm]{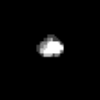} &
        \includegraphics[width=1.35cm]{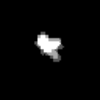} &
        \includegraphics[width=1.35cm]{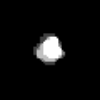} &
        \includegraphics[width=1.35cm]{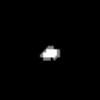} &
        \includegraphics[width=1.35cm]{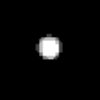} &
        \includegraphics[width=1.35cm]{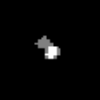} &
        \includegraphics[width=1.35cm]{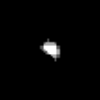} \\
        \hline \\[-0.8em]
        \small\rotatebox{90}{\parbox[c][0.2cm][c]{1.35cm}{Standard}} \hspace{3pt}
        \includegraphics[width=1.35cm]{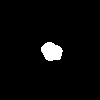} &
        \includegraphics[width=1.35cm]{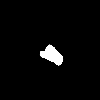} &
        \includegraphics[width=1.35cm]{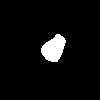} &
        \includegraphics[width=1.35cm]{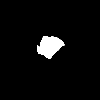} &
        \includegraphics[width=1.35cm]{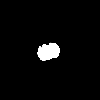} &
        \includegraphics[width=1.35cm]{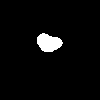} &
        \includegraphics[width=1.35cm]{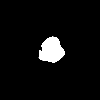} &
        \includegraphics[width=1.35cm]{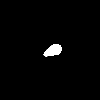} &
        \includegraphics[width=1.35cm]{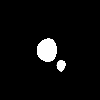} &
        \includegraphics[width=1.35cm]{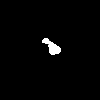} &
        \includegraphics[width=1.35cm]{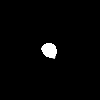} \\
        \small\rotatebox{90}{\parbox[c][0.2cm][c]{1.35cm}{CC}} \hspace{3pt}
        \includegraphics[width=1.35cm]{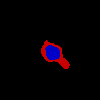} &
        \includegraphics[width=1.35cm]{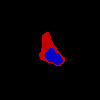} &
        \includegraphics[width=1.35cm]{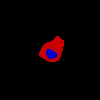} &
        \includegraphics[width=1.35cm]{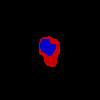} &
        \includegraphics[width=1.35cm]{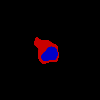} &
        \includegraphics[width=1.35cm]{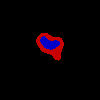} &
        \includegraphics[width=1.35cm]{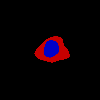} &
        \includegraphics[width=1.35cm]{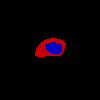} &
        \includegraphics[width=1.35cm]{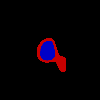} &
        \includegraphics[width=1.35cm]{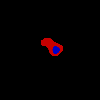} &
        \includegraphics[width=1.35cm]{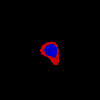}\\
        \small\rotatebox{90}{\parbox[c][0.2cm][c]{1.35cm}{Bayesian}} \hspace{3pt}
        \includegraphics[width=1.35cm]{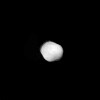} &
        \includegraphics[width=1.35cm]{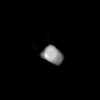} &
        \includegraphics[width=1.35cm]{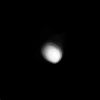} &
        \includegraphics[width=1.35cm]{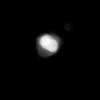} &
        \includegraphics[width=1.35cm]{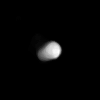} &
        \includegraphics[width=1.35cm]{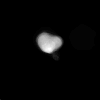} &
        \includegraphics[width=1.35cm]{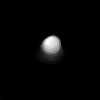} &
        \includegraphics[width=1.35cm]{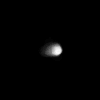} &
        \includegraphics[width=1.35cm]{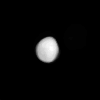} &
        \includegraphics[width=1.35cm]{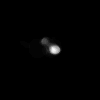} &
        \includegraphics[width=1.35cm]{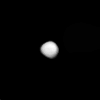}
    \end{tabular}
    \setlength\tabcolsep{6pt}
    \caption{Examples from LIDC. \textbf{Image}: the original image \textit{data}; \textbf{Ground}: original annotation \textit{data} from LIDC, composited such that a brighter pixel indicates that more annotators marked it in the positive class; \textbf{Standard}: singular \textit{predictions} from an attention U-Net trained on original LIDC annotations; \textbf{CC}: CC-style \textit{predictions} of an attention U-Net trained on our Confidence Contour annotations, (blue: min contour, red: max contour); \textbf{Bayesian}: the uncertainty map from a Bayesian attention U-Net with test-time dropout. 
    We include the uncertainty map from Bayesian U-Net as an example of continuous uncertainty maps to illustrate interpretability differences compared to CC-style predictions.
    }
    \label{tab:largepanel}
\end{figure*}


We investigate whether general segmentation models can successfully model CC annotations. 
To adapt an existing general segmentation model to learn CCs, we can map CC labels into two-channel masks. A model trained on these masks conceptually behaves as two segmentation models with heavily shared weights, which predict the min (`min-subnetwork') and the max (`max-subnetwork') contours separately. 
Each individual subnetwork is formally equivalent in structure to a general model trained on singular-type annotations.
In our modeling experiments, we observe no statistically significant difference ($p > 0.05$) between the converged loss of singular annotations and either the `min-subnetwork' or the `max-subnetwork', across all 144 ($48 + 48 + 12 + 36$) modeling trials.
This shows that it is not more difficult for a wide range of general segmentation models to learn to predict CC labels than singular ones.

While we observe no performance losses when training on CCs or similar labels, we experience significant improvements in the interpretability of the uncertainty predictions. Figure~\ref{tab:largepanel} visualizes predictions from different models across multiple samples from LIDC. 
Rather than predicting singular annotations, which do not provide explicit information about uncertainty, models trained on CCs explicitly report areas of high and low annotator confidence.
Moreover, as opposed to previously mentioned model-based uncertainty map generation methods such as Bayesian segmentation models which produce smooth, unthresholded maps, our models' discrete predictions provide clearly interpretable uncertainty thresholds directly corresponding to human annotations.
We note that while alternative approaches such as elicitation and candidate generation produce uncertainty representations through aggregating multiple singular predictions (a kind of ``disagreement'' uncertainty), using CCs reproduces uncertainty as assessed by a single annotator (a kind of ``ambiguity'' uncertainty).

\subsection{Expert Impressions of Interpretability and Utility}
\subsubsection{Method}
We recruited five experts with backgrounds in medical imaging for thirty-minute interviews (Table 4). First, we asked the experts to look at annotations from 6--9 (depending on time) high-disagreement samples in the LIDC dataset, explain how they interpreted the disagreement, and show how they would have annotated the image. Second, we showed annotators 5 samples along with three model predictions---standard (singular binary mask), CC, and Bayesian---and the original image without additional context (similarly to what is shown in Figure~\ref{tab:largepanel}). Experts were asked to interpret what the different masks meant and which they felt best corresponded to the original image. Last, we explained how each of the three model predictions were derived and asked experts to reflect on how they might use each of the models in practice, with the ability to flip through 15 additional samples. Each expert was compensated \$50.

Throughout the analysis, we use `Bayesian' when the expert is talking about a feature of the predictions which is dependent on the specific method of Bayesian segmentation, and `continuous maps' when the expert is talking about continuity of uncertainty values, which is a feature of many alternative uncertainty representation methods. Therefore, statements on continuous maps can be generalized to all approaches which present uncertainty in a similarly unthresholded manner.

\begin{table*}
    \centering
    \begin{tabular}{c|l|p{100mm}|c}
        \toprule
        ID & Degree & Position and Background & Worked with AI? \\
        \midrule
        P1 & Ph.D. student & Computer Science, medical image segmentation for pathology & \cmark \\ 
        P2 & M.D. & Research scientist in pathology & \xmark \\
        P3 & M.D. & Doctor at ER and other clinics, experienced in lung X-rays & \xmark \\
        P4 & M.D. & Radiation oncology resident, some work in medical segmentation tools & \cmark \\
        P5 & M.D., Ph.D. & Radiology resident, medical imaging ML and UI research & \cmark \\
        \bottomrule
    \end{tabular}
    \caption{Expert information table. ``Worked with AI'' means familiarity not only with how to interact with AI models, but their internal structure and behavior, as evidenced by publications building medical AI systems.}
\end{table*}




\subsubsection{Max contours in CCs encompass higher-sensitivity information compared to singular annotations.}
Experts noticed that the max contour of Confidence Contours annotations tended to pick up on higher-sensitivity information than singular annotations. In several cases, an ambiguous protrusion was connected to a high-certainty mass; while both the standard and the Bayesian predictions neglected the ambiguous area, the max contour of the CC prediction encapsulated it. 
P1 made the accurate observation that ``\textit{the ground truth of the mask can only give you so much... this [singular annotation] is probably what the human labeler is labeling for [continuous annotations] anyway}.'' 
That is, Bayesian maps are a reformulation of the same data and model used to produce singular masks.
P2, who was less trustful of models and almost always preferred to work with the images directly, said that they preferred Confidence Contours out of all three possible masks because it gave them the most information.

\subsubsection{CCs are particularly useful when \textit{structural} uncertainty is present and relevant to the task.}
By `structural uncertainty,' we mean that the uncertainty does not stem strictly from blurriness, stains, or other visual impediments, but rather concerns a concrete structure in the image. This structure may be uncertain in part because of visual impediments, but also due to its relationships with other structures (which requires domain knowledge to identify). Structural uncertainty is often demonstrated by discrete disagreement between singular annotations, in which annotators either include or exclude a particular structure with no `in between.'

Experts identified structural uncertainty in many LIDC samples and explained differences between singular annotations as experts following different reasoning schemes rather than perception-level differences (``\textit{she just happened to see X or adopt Y sensitivity level}''). Across the experts, we did not observe consensus on the best annotations.
Confidence Contours works well in these cases, because uncertainty is built into the structure of the problem and needs to be accounted for in measurements in a clear way.
P5 argued that ``\textit{[CCs] is going to be advantageous in situations where there are clustered nodules or a solid ground-glass nodule that has more of a halo which also needs to be measured.}''
Moreover, experts' ability to rationalize other experts' annotations provides further support that experts are able to create min and max contours which accurately bound the range of uncertainty.

However, when an image features low structural uncertainty, annotators are content with singular annotation masks because they provide all the information that is needed. 
P1 noted that ``\textit{Just doing a binary segmentation will give you a very crude estimate, but it's the most easy to just identify.}''
P5 concurs: ``\textit{[its] benefit is in the simplicity}.''
Moreover, experts note that singular annotations suffice when the expert only needs to measure features in which disagreement from structural uncertainty tends not to matter, such as the rough location of a nodule. 

\subsubsection{Making judgements using continuous maps is more challenging due to the lack of a threshold and decay of edges.}
In contrast, experts stated that the unthresholded nature of continuous maps makes it difficult to efficiently and reliably make judgements.
This aligns with the challenges expressed in the literature discussed earlier.
P1 commented: ``\textit{The problem with [continuous maps] if I were looking at it just with my eye is that it's really difficult to tell the certainty level... it would be nice to have some range or threshold}''. 
On the other hand, with Confidence Contours, ``\textit{You can get a really clean outline of the node, and you also have some idea of the uncertainty}.'' 
P4 remarked that ``\textit{[continuous maps] could be helpful, but you have to check the color scales, look at the numbers, so it would be slower... if I had to pick one [segmentation map], it is probably [CCs].'' }
P4 also noted later that calibrating color scales across samples and ensuring that they are comparable would be another difficulty of practically using continuous maps.
P5 observed, ``\textit{It's cool that [continuous maps] are more of a continuous metric of what is in the nodule and what is not in the nodule, but for conventional clinical reporting, it's difficult to know where to draw the line, which makes it kind of hard to use.}''
On the other hand, the double-thresholded nature of CCs often more intuitively captures the distribution of uncertainty in such problems.

 Experts also noted that continuous maps can obscure the shape of the region of interest, as the more ambiguous but significant edges may be decayed. 
P5 brought up that the Bayesian approach makes it less easy to see what is going on, noting that ``\textit{[Singular annotations] and [Confidence Contours] do a pretty good job of showing---here is a shape, and here is the more solid area in the shape... so I probably prefer [singular annotations] and [CCs] over [continuous masks].}''
P3, who was very focused on the shapes of the nodes, disliked continuous maps because it does not clearly present this important discretized feature.
On the other hand, the thresholded nature of both singular and CCs annotations enables a more definite presentation of the nodule shape.


\subsubsection{CCs provide interpretable boundaries enabling greater expert choice, while continuous maps can be misinterpreted.}

Some experts found the two-level sensitivity representation method more `natural' when comparing it to their own process of interpretation of an image.
P5 mentioned that ``\textit{you may get into situations where you report two separate measurements, a solid measurement and a ground glass measurement. [CCs] makes it easier to automate that.}''  We further explore this in the Discussion section (``Greater representation of uncertainty range'').

P5 also commented that ``\textit{[CCs] is easier to understand because I feel like the [min] contour is something which is more reliable that you can fall back on, and you can use the [max] contour if it makes sense to, given the situation. But you don't get those two channels of information in [continuous maps].}'' That is, when making a judgement, one can begin from the min contour and move outwards as necessary. 
For instance, P3---who we observed tended to draw very strict boundaries at low sensitivity---said that CCs segmentations most reflected the information in the original image because it gave them the choice to focus on the min contour and to ignore the max contour. 
In this way, the expert has more control over their judgement of the image due to the intepretability of the contours, whereas in singular annotations they are given no alternatives and in continuous maps little guidance.




Indeed, continuous maps can lead experts without a background in AI astray.
We noticed that P3, who was not familiar with AI, misinterpreted continuous maps as corresponding to intensity levels (brightness) in the original image.
While for lung nodule segmentation, intensity does correlate well with certainty, they are different concepts: notably, a pixel/voxel can have low intensity but high certainty, due to its spatial relations to other structures. Experts familiar with AI were able to speak about these two concepts separately.


\section{Discussion}
\label{place:discussion}

\paragraph{Taking a data-centric approach.} Broadly, our work provides a data-centric supplement to the dominantly model-centric work in uncertainty representation in semantic segmentation. 
We show that using data with explicit uncertainty markings to train general models can directly produce more interpretable uncertainty maps than training complex models fitted with generative or probabilistic components on singular annotations, without loss of performance.
In deployment settings, it may be more feasible to adopt such an approach to minimize the burden of infrastructure modification while producing more diverse functionality.

\paragraph{Distinguishing sources of uncertainty.}
When annotators use CCs over singular annotations, more of the uncertainty in the dataset is accounted for directly within the structure and less of it is present in disagreements between annotators. 
This is suggested by our results finding that the min and max contour have statistically significantly reduced cross-annotator disagreement than singular annotations, across all datasets. 
Remaining disagreements between annotations are more likely to result from irreducible `core' annotator disagreements~\cite{Kairam2016PartingCC} such as perception or medical background~\cite{schaekermann_eeg_adjudication_2019}.

\paragraph{For uncertainty, more isn't necessarily better.} 
All experts in our interview generally preferred singular or Confidence Contours annotations, depending on the presence of structural uncertainty. 
Continuous maps, on the other hand, end up producing ``uncertainty about uncertainty'' because experts are unsure how to formulate a judgement from the collection of uncertainty information. Showing the raw distribution of probabilities can increase mental load, whereas choosing the right presentation can improve understanding, even if less absolute information is being shown~\cite{Padilla2021UncertainAU}.
Moreover, as discussed in the interview results, it may also lead to incorrect interpretations.
Despite being well-documented in the field of human computation, it seems that most of the work in uncertainty-aware segmentation does not consider these human judgement limitations due to its model-centric focus.

\paragraph{Greater representation of uncertainty range.} 
We hypothesize that Confidence Contours (CCs) enable annotators to map wider regions of uncertainty than singular annotations, which in turn allows models to access explicit learning signals in such ambiguous areas.
This is suggested by our observation that the average max annotation is larger than the average singular annotation (+25.6\% LIDC and  +17.0\% FoggyBlob) coupled with the low overflow we see in ``Evaluating Representative Capacity''.
Model-centric approaches to uncertainty representation can be broadly conceptualized as inferring a distribution of singular annotations from $k$ sampled singular annotations for each image.
Dominant methods in the field are such a case with $k=1$. Alternative approaches that attempt to train models on labels derived from multiple annotators' singular annotations per image~\cite{Hu2019SupervisedUQ,Fornaciari2021BeyondB} use higher values of $k$.
Such model-centric approaches, then, still all receive positive learning signals only from labels produced near a high-certainty threshold and only represent ``sufficiently certain'' uncertainty.
CCs, however, allow annotators to directly communicate both a threshold for what is considered ``certain'' and a threshold for what is considered ``possible''. 
In our interviews, we found that medical experts appreciated having a wider range of signals to provide more information for judgement-making. Moreover, by representing a larger range of uncertainty, CCs are able to accommodate a wide range of annotator preferences; we observed that experts with both low- and high-sensitivity tendencies were able to map their preferred segmentations to CCs more than to singular masks.



\paragraph{CC-compatible models from existing datasets.}
\begin{figure*}[!t]
    \centering
    \setlength\tabcolsep{1.9pt}
    \begin{tabular}{c ccccc ccccc c}
        \small\rotatebox{90}{\parbox[c][0.2cm][c]{1.35cm}{CC}} \hspace{3pt}
        \includegraphics[width=1.35cm]{imgs/34.png} &
        \includegraphics[width=1.35cm]{imgs/35.png} &
        \includegraphics[width=1.35cm]{imgs/36.png} &
        \includegraphics[width=1.35cm]{imgs/37.png} &
        \includegraphics[width=1.35cm]{imgs/38.png} &
        \includegraphics[width=1.35cm]{imgs/39.png} &
        \includegraphics[width=1.35cm]{imgs/40.png} &
        \includegraphics[width=1.35cm]{imgs/41.png} &
        \includegraphics[width=1.35cm]{imgs/42.png} &
        \includegraphics[width=1.35cm]{imgs/43.png} &
        \includegraphics[width=1.35cm]{imgs/44.png}\\
        \small\rotatebox{90}{\parbox[c][0.2cm][c]{1.35cm}{PCC}} \hspace{3pt}
        \includegraphics[width=1.35cm]{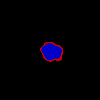} &
        \includegraphics[width=1.35cm]{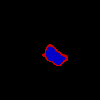} &
        \includegraphics[width=1.35cm]{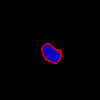} &
        \includegraphics[width=1.35cm]{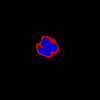} &
        \includegraphics[width=1.35cm]{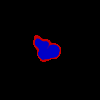} &
        \includegraphics[width=1.35cm]{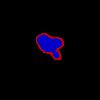} &
        \includegraphics[width=1.35cm]{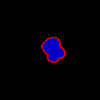} &
        \includegraphics[width=1.35cm]{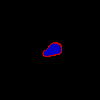} &
        \includegraphics[width=1.35cm]{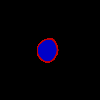} &
        \includegraphics[width=1.35cm]{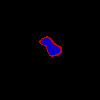} &
        \includegraphics[width=1.35cm]{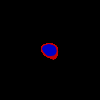} \\
    \end{tabular}
    \setlength\tabcolsep{6pt}
    \caption{Examples from LIDC, compared between \textbf{CC} (Confidence Contours) and \textbf{PCC}: CC-style \textit{predictions} of an attention U-Net trained on pseudo-CC data inferred from ground LIDC annotations (Section~\ref{place:discussion}).
    }
    \label{tab:smallpanel}
\end{figure*}
We have observed that models predicting CCs have substantive interpretability benefits, but it may be costly to re-annotate existing datasets with CCs. Can we approximate some of these interpretability benefits before or without collecting CCs? We also explore the inference of high- and low-confidence regions from existing segmentation datasets with disaggregated annotations. For a given set $S$ of singular annotations, we can define the intersection of all annotations as an approximate `min' contour and the union as an approximate `max' contour.
$$\hat{y}_\text{min} = \bigcap_{s \in S} s, \hat{y}_\text{max} = \bigcup_{s \in S} s$$
In experiments with the same setup as ``Modeling Viability'', we find that general segmentation models trained on these `intersection-union'-style labels perform as well in training and validation performance as models trained on aggregated labels. 
In this approach, disagreement between annotations is used as a proxy for uncertainty.
Figure~\ref{tab:smallpanel} compares CCs against these CC-approximations. Although both have the same two-level representation, we can see that the CC-approximations have a much smaller max contour size.
This is because CC-approximating model predictions are not able to capture structural uncertainty as well as CCs, as they are derived from singular annotations with limited structural uncertainty information. The max contour in these cases serves as a relatively uniform bound on purely \textit{visual} uncertainty (blurring, staining, etc.).
As such, we caution against using these approximated masks to replace directly-provided annotations of uncertainty.
However, intersection/union predictions still benefit from clearer interpretability: the thresholding provided by CC-like representations can reduce  cognitive burden compared to understanding uncertainty maps and candidate ensembles.

\paragraph{Limitations and future work.} 
In order to explore Confidence Contours in a well-studied and relatively unchallenging context, we evaluated on lung nodule segmentation.
While lung nodule images often exhibit at least some form of structural uncertainty, in practice treatments are conducted with added safety margins which encompass most of this uncertainty. Moreover, while shape features are important in particular cases, radiologists are often more interested in the size and location of the nodule. In these contexts, it may be more effort for the radiologist to verify that the model is correct than for the radiologist to estimate the ground truth themselves, which makes it ill-suited for modeling~\cite{Fok2023InSO}. However, the medical experts in our interviews proposed the brain and the abdomen as regions in which a) have much less clear delineations between structures and therefore require the assistance of models, and b) are areas in which highly specific operations which require a clear mapping of structure shapes and low margins are performed. Future work would evaluate data-centric uncertainty approaches in these domains, where the clinical significance of using Confidence Contours can be clearer.

Another area of future work would be to probe for data-centric uncertainty modeling as an explainable AI (XAI) method. Existing uncertainty modeling methods such as Bayesian segmentation can be interpreted both as showing the distribution of uncertainty in the image, and providing a visual explanation for the learned representations and behavior of the model itself. It may be, for instance, that the \textit{models} which are trained on Confidence Contours not only produce more interpretable uncertainty maps, but are themselves more interpretable.



\section{Conclusion}
Medical semantic segmentation is a particularly human-involved application of modeling: domain experts annotate the labels that models are trained on, and these models will eventually produce predictions which will be used by medical decision-makers for patients. 
In this pipeline, it is important not only to develop powerful models but to ensure that such models are trained on data that effectively captures uncertainty and that produce usable and informative predictions.
Our work explored how adopting a data-centric approach can better accommodate the people on both ends of the modeling pipeline---the annotators and the judgement-makers.
We proposed Confidence Contours as an novel segmentation annotation representation which explicitly and effectively marks uncertainty. 
Confidence Contour annotations can be used with general models to produce highly interpretable and practically useful uncertainty maps without loss of performance.

\section*{Ethical Statement}
Existing work on uncertainty estimation in medical segmentation produces uncertainty maps or candidates which are difficult to interpret. Our work moves towards uncertainty-reporting models whose uncertainty maps are easy to interpret to humans. This may aid in understanding model responsibility and trustworthiness when making jointly informed medical decisions.



\bibliography{aaai23,extra}

\end{document}